\documentclass[english]{IEEEtran}
\usepackage[T1]{fontenc}
\usepackage[utf8]{inputenc}
\usepackage{color}
\usepackage{url}
\usepackage{amsmath}
\usepackage{amssymb}
\usepackage{stmaryrd}
\usepackage{graphicx}

\makeatletter

\providecommand{\tabularnewline}{\\}

\usepackage{stmaryrd}
\usepackage[utf8]{inputenc}

\@ifundefined{showcaptionsetup}{}{%
 \PassOptionsToPackage{caption=false}{subfig}}
\usepackage{subfig}
\makeatother

\usepackage{babel}
\begin{document}

\title{$t$-Exponential Memory Networks for Question-Answering Machines}

\author{Kyriakos Tolias and Sotirios P. Chatzis\thanks{The authors are with the Department of Electrical Engineering, Computer
Engineering, and Informatics, Cyprus University of Technology, Cyprus.}}
\maketitle
\begin{abstract}
Recent advances in deep learning have brought to the fore models that
can make multiple computational steps in the service of completing
a task; these are capable of describing long-term dependencies in
sequential data. Novel recurrent attention models over possibly large
external memory modules constitute the core mechanisms that enable
these capabilities. Our work addresses learning subtler and more complex
underlying temporal dynamics in language modeling tasks that deal
with sparse sequential data. To this end, we improve upon these recent
advances, by adopting concepts from the field of Bayesian statistics,
namely variational inference. Our proposed approach consists in treating
the network parameters as latent variables with a prior distribution
imposed over them. Our statistical assumptions go beyond the standard
practice of postulating Gaussian priors. Indeed, to allow for handling
outliers, which are prevalent in long observed sequences of multivariate
data, multivariate $t$-exponential distributions are imposed. On
this basis, we proceed to infer corresponding posteriors; these can
be used for inference and prediction at test time, in a way that accounts
for the uncertainty in the available sparse training data. Specifically,
to allow for our approach to best exploit the merits of the $t$-exponential
family, our method considers a new $t$-divergence measure, which
generalizes the concept of the Kullback-Leibler divergence. We perform
an extensive experimental evaluation of our approach, using challenging
language modeling benchmarks, and illustrate its superiority over
existing state-of-the-art techniques.
\end{abstract}


\section{Introduction}

Recent developments in machine learning have managed to achieve breakthrough
improvements in modeling long-term dependencies in sequential data.
Specifically, the machine learning community has recently witnessed
a resurgence in models of computation that use explicit storage and
a notion of attention \cite{turing,mem,n2n-mem,sum}. As it has been
extensively shown, the capability of effectively manipulating such
storage mechanisms offers a very potent solution to the problem of
modeling long temporal dependencies. Its advantages have been particularly
profound in the context of question-answering bots. In such applications,
it is required that the trained models be capable of taking multiple
computational steps in the service of answering a question or completing
a related task. 

This work builds upon these developments, seeking novel treatments
of Memory Networks (MEM-NNs) \cite{mem,n2n-mem} to allow for more
flexible and effective learning from sparse sequential data with heavy-tailed
underlying densities. Indeed, both sparsity and heavy tails are salient
characteristics in a large variety of real-world language modeling
tasks. Specifically, the earliest solid empirical evidence that any
sufficiently large corpus of natural language utterances entails heavy-tailed
distributions with power-law nature dates back to 1935 \cite{zipf}.
Hence, we posit that the capability of better addressing these data
properties might allow for advancing the state-of-the-art in the field.
Our inspiration is drawn from recent developments in approximate Bayesian
inference for deep learning models \cite{aevb,aevb2,aevb3,aevb4,vb-weights}.
Bayesian inference in the context of deep learning models can be performed
by considering that the network parameters are stochastic latent variables
with some prior distribution imposed over them. This inferential framework
allows for the developed network to account for the uncertainty in
the available (sparse) training data. Thus, it is expected to yield
improved predictive and inferential performance outcomes compared
to the alternatives. 

Existing Bayesian inference formulations of deep networks postulate
Gaussian assumptions regarding the form of the imposed priors and
corresponding (inferred) posterior distributions. Then, inference
can be performed in an approximate, computationally efficient way,
by resorting to variational Bayes \cite{vbg}. This consists in searching
for a proxy in an analytically solvable distribution family that approximates
the true underlying distribution. To measure the closeness between
the true and the approximate distribution, the relative entropy between
these two distributions is used. Specifically, under the aforementioned
Gaussian assumption, one can use the Shannon-Boltzmann-Gibbs (SBG)
entropy, whereby the relative entropy yields the well known Kullback-Leibler
(KL) divergence \cite{entropy}. 

Despite these advances, in problems dealing with long sequential data
comprising multivariate observations, such assumptions of normality
are expected to be far from the actual underlying densities. Indeed,
it is well-known that real-world multivariate sequential observations
tend to entail a great deal of outliers (heavy-tailed nature). This
fact gives rise to significant difficulties in data modeling, the
immensity of which increases with the dimensionality of the data \cite{key-30}.
Hence, replacing the typical Gaussian assumption with alternatives
has been recently proposed as a solution towards the amelioration
of these issues \cite{asydgm}. 

Our work focuses on the $t$-exponential family\footnote{Also referred to as the $q$-exponential family or the Tsallis distribution.},
which was first proposed by Tsallis and co-workers \cite{tsalis,tsalis2,tsalis3},
and constitutes a special case of the more general $\phi$-exponential
family \cite{naudts,naudts2,naudts3}. Of specific practical interest
to us is the Students'-$t$ density, which has been extensively examined
in the literature of generative models, such as hidden Markov models
\cite{vbshmm,vbtmfa,shmm}. The Student's-$t$ distribution is a bell-shaped
distribution with heavier tails and one more parameter (degrees of
freedom - DOF) compared to the normal distribution, and tends to a
normal distribution for large DOF values \cite{key-19}. Hence, it
provides a much more robust approach to the fitting of models with
Gaussian assumptions. On top of these merits, the $t$-exponential
family also gives rise to a new $t$-divergence measure; this can
be used for performing variational inference in a fashion that better
accommodates heavy-tailed data (compared to standard KL-based solutions)
\cite{tVB}. 

Under this rationale, our proposed approach is founded upon the fundamental
assumption that the imposed priors over the postulated MEM-NN parameters
are Student's-$t$ densities. On this basis, we proceed to infer their
corresponding Student's-$t$ posteriors, using the available training
data. To best exploit the merits of the $t$-exponential family, we
effect variational inference by resorting to a novel algorithm formulation;
this consists in minimizing the $t$-divergence measure \cite{tVB}
over the sought family of approximate posteriors. 

The contribution of this work can be summarized as follows:
\begin{enumerate}
\item The proposed approach, dubbed $t$-exponential Memory Network ($t-$MEM-NN),
is the first ever attempt to derive a Bayesian inference treatment
of MEM-NN models for question-answering. Our approach imposes a prior
distribution over model parameters, and obtains a corresponding posterior;
this is in contrast to existing approaches, which train simple point-estimates
of the model parameters. By obtaining a full posterior density, as
opposed to a single point-estimate of the model parameters, our approach
is capable of coping with uncertainty in the trained model parameters. 
\item We consider imposition of Student's-$t$ priors, which are more appropriate
for applications dealing with modeling heavy-tailed phenomena, as
is the case with large natural language corpora \cite{zipf}. This
is the first time that explicit heavy-tailed distribution modeling
is considered in the literature of MEM-NNs. Eventually, by making
use of the trained posteriors, one can perform inference by drawing
multiple alternative samples of the model parameters, and averaging
the predictive outcomes pertaining to each sample. Thus, our predictions
do not rely on the ``correctness'' of just a single model estimate;
this way, the effects of model uncertainty are considerably ameliorated.
\item Model training is performed by maximizing a $t$-divergence-based
objective functional, as opposed to the commonly used objectives that
are based on the KL divergence. This allows for making the most out
of the heavy tails of the obtained Student's-$t$ distributions. Our
work is the first one that performs approximate inference for deep
latent variable models on the grounds of a $t$-divergence-based objective
functional.
\end{enumerate}
The remainder of this paper is organized as follows: In the following
Section, we provide a brief overview of the related work. In Section
III, our approach is introduced; specifically, we elaborate on its
motivation, formally define our proposed model, and derive its training
and inference algorithms. In Sections IV and V, we perform the experimental
evaluation of our approach, and illustrate its merits over the current
state-of-the-art. Finally, the concluding Section of this paper summarizes
our contribution and discusses our results.

\section{Methodological Background}

\subsection{End-to-end Memory Networks}

Our proposed approach extends upon the existing theory of MEM-NNs,
first introduced in \cite{mem}. Specifically, we are interested in
a recent end-to-end-trainable extension of MEM-NNs, presented in \cite{n2n-mem}.
That variant enjoys the advantage of requiring much less supervision
during training, which is of major importance in real-world question-answering
scenarios. The model input comprises a set of \emph{facts, $\{\boldsymbol{x}_{i}\}_{i=1}^{N}$,}
that are to be stored in the memory, as well as a \emph{query} $\boldsymbol{q}$;
given these, the model outputs an \emph{answer} $a$. Each of the
facts\emph{, $\boldsymbol{x}_{i}$, }as well as the query, $\boldsymbol{q}$,
contain symbols coming from a dictionary with $V$ words. Specifically,
they are represented by vectors that are computed by concatenating
the one-hot representations of the words they contain. The latter
are obtained on the basis of the available dictionary comprising $V$
words. The model writes all $\boldsymbol{x}_{i}$ to the memory, up
to a fixed buffer size, and then finds a continuous variable encoding
for both the $\boldsymbol{x}_{i}$ and the $\boldsymbol{q}$. These
continuous representations are then processed via multiple hops, so
as to generate the output $a$; this essentially constitutes one (selected)
symbol from the available dictionary. This modeling scheme allows
for establishing a potent training procedure, which can perform multiple
memory accesses back to the input.

Specifically, let us consider one layer of the MEM-NN model. This
is capable of performing a single memory hop operation; multiple hops
in memory can be performed by simply stacking multiple such layers\footnote{The number of hops performed in memory is a model hyperparameter,
that has to be selected in a heuristic manner. Naturally, there is
no point in this number exceeding the number of facts presented to
the model each time.}. It comprises three main functional components:\\
\textbf{Input memory representation}: Let us consider an input set
of facts\emph{, $\{\boldsymbol{x}_{i}\}_{i=1}^{N}$,} to be stored
in memory. This entire set is converted into \emph{memory vectors},\emph{
$\{\boldsymbol{m}_{i}\}_{i=1}^{N}$, $\boldsymbol{m}_{i}\in\mathbb{R}^{\delta}$,}
computed by embedding each $\boldsymbol{x}_{i}$ in a continuous space,
using a position embedding procedure \cite{n2n-mem} with embedding
matrix $\boldsymbol{A}$. The query $\boldsymbol{q}$ is also embedded
in the same space; this is performed via a position embedding procedure
with embedding matrix $\boldsymbol{B}$, and yields an \emph{internal
state vector }$\boldsymbol{u}$. On this basis, MEM-NN proceeds to
compute the match between the submitted query, $\boldsymbol{q}$,
and each one of the available facts, by exploiting the salient information
contained in their inferred embeddings; that is, the state vector,
$\boldsymbol{u}$, and the memory vectors,\emph{ $\{\boldsymbol{m}_{i}\}_{i=1}^{N}$,
}respectively. Specifically, it simply takes their inner product followed
by a softmax:
\begin{equation}
\varpi_{i}=\mathrm{softmax}(\boldsymbol{u}^{T}\boldsymbol{m}_{i})
\end{equation}
where 
\begin{equation}
\mathrm{softmax}(\boldsymbol{\zeta}_{i})\triangleq\frac{\mathrm{exp}(\boldsymbol{\zeta}_{i})}{\sum_{j}\mathrm{exp}(\boldsymbol{\zeta}_{j})}
\end{equation}
In essence, $\boldsymbol{\varpi}=[\varpi_{i}]_{i=1}^{N}$, is a probability
vector over the facts, which shows how strong their affinity is with
the submitted query. We will be referring to this vector as the inferred
\emph{attention }vector.\\
\textbf{Output memory representation}: In addition to the inferred
memory vectors, MEM-NN also extracts from each fact, $\boldsymbol{x}_{i}$,
a corresponding \emph{output vector embedding,} $\boldsymbol{c}_{i}$,
via another position embedding procedure \cite{n2n-mem} with embedding
matrix $\boldsymbol{C}$. These output vector embeddings are considered
to encode the salient information included in the presented facts
that can be used for output (answer) generation. To achieve this goal,
we leverage the inferred attention vector $\boldsymbol{\varpi},$
by using it to weight each fact (encoded via its inferred output vector
embedding) with the corresponding computed probability value. It holds
\begin{equation}
\boldsymbol{o}=\sum_{i=1}^{N}\varpi_{i}\boldsymbol{c}_{i}
\end{equation}
\\
\textbf{Generating the final prediction}: MEM-NN output layer is a
simple softmax layer, which is presented with the computed output
vector, $\boldsymbol{o}$, as well as the internal state vector, $\boldsymbol{u}$.
It estimates a probability vector over all possible predictions, $\hat{\boldsymbol{a}}$,
that is all the entries of the considered dictionary of size $V$.
It holds
\begin{equation}
\hat{\boldsymbol{a}}=\mathrm{softmax}(\boldsymbol{W}(\boldsymbol{o}+\boldsymbol{u}))
\end{equation}
 where $\boldsymbol{W}$ is the weights matrix of the output layer
of the network, whereby we postulate
\[
a=\mathrm{arg\,max}(\hat{\boldsymbol{a}})
\]

A graphical illustration of the considered end–to-end trainable MEM-NN
model, that we build upon in this work, is provided in Fig. 1. Our
exhibition includes both single-layer models, capable of performing
single memory hop operations, as well as multi-layer ones, obtained
by stacking multiple singe layers, which can perform multiple hops
in memory. Note that, to save parameters, and reduce the model's overfitting
tendency, as well as its memory footprint, we tie the corresponding
embedding matrices across all MEM-NN layers, as suggested in \cite{n2n-mem}.

\begin{figure*}
\begin{centering}
\includegraphics[scale=0.25]{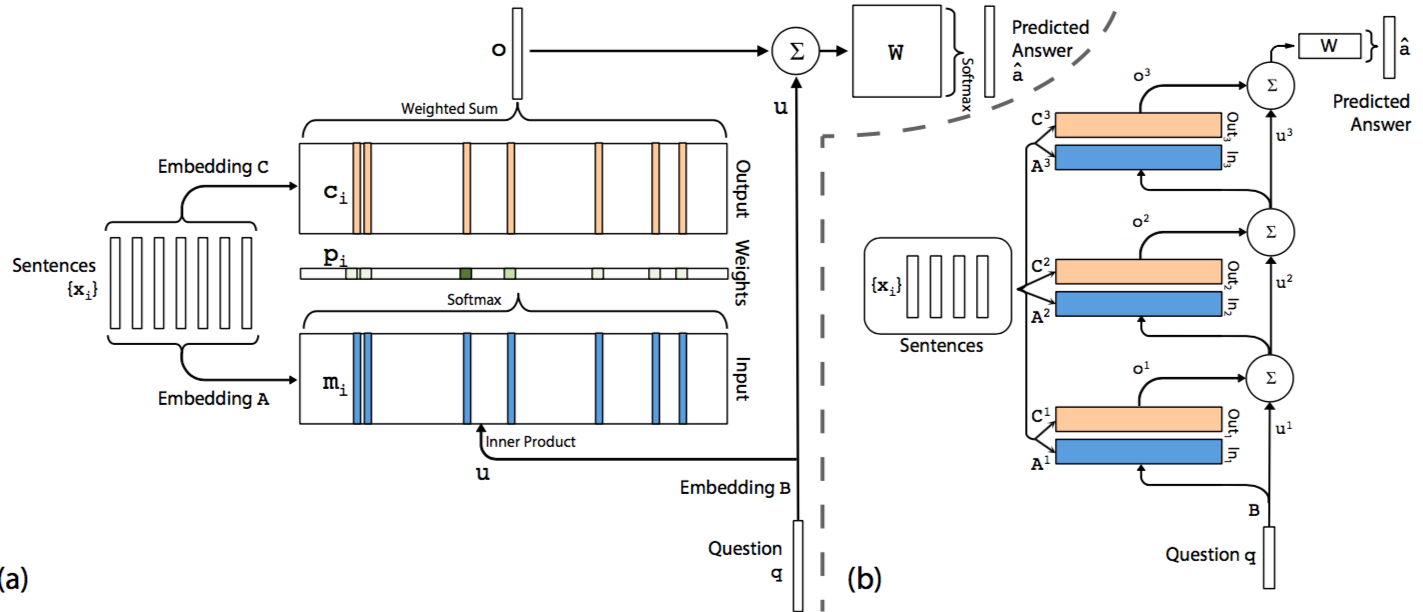}
\par\end{centering}
\caption{(a) A single-layer version of the considered model. (b) A 3-layer
version, obtained via stacking (adopted from \cite{n2n-mem}).}

\end{figure*}

\subsection{Variational Bayes in Deep Learning}

The main idea of applying variational Bayesian inference to deep learning
models consists in calculating a posterior distribution over the network
weights given the training data. The benefit of such a learning algorithm
setup is that the so-obtained posterior distribution answers predictive
queries about unseen data by taking expectations: Prediction is made
by averaging the resulting predictions for each possible configuration
of the weights, weighted according to their posterior distribution.
This allows for accounting for uncertainty, which is prevalent in
tasks dealing with sparse training datasets.

Specifically, let us consider a training dataset $\mathcal{D}$. A
deep network essentially postulates and fits to the training data
a (conditional) likelihood function of the form $p(\mathcal{D}|\boldsymbol{w})$,
where $\boldsymbol{w}$ is the vector of network weights. In the case
of Bayesian treatments of neural networks, an appropriate prior distribution,
$p(\boldsymbol{w})$, is imposed over $\boldsymbol{w}$, and the corresponding
posterior is inferred from the data \cite{vb-weights}. This consists
in introducing an approximate posterior distribution over the network
weights, $q(\boldsymbol{w};\boldsymbol{\phi})$, and optimizing it
w.r.t. a lower bound to the network log-marginal likelihood $\mathrm{log}\,p(\mathcal{D};\boldsymbol{\phi})$,
commonly referred to as the evidence lower bound (ELBO), $\mathcal{L}(\boldsymbol{\phi})$
\cite{attias}; it holds
\begin{equation}
\begin{aligned}\mathrm{log}\,p(\mathcal{D};\boldsymbol{\phi})\geq\mathcal{L}(\boldsymbol{\phi})= & \mathbb{E}_{q(\boldsymbol{w};\boldsymbol{\phi})}[\mathrm{log}\,p(\mathcal{D}|\boldsymbol{w})\\
 & +\mathrm{log}\,p(\boldsymbol{w})-\mathrm{log}\,q(\boldsymbol{w};\boldsymbol{\phi})]
\end{aligned}
\end{equation}
where $\mathbb{E}_{q(\boldsymbol{w};\boldsymbol{\phi})}[\cdot]$ is
the expectation of a function w.r.t. the random variable $\boldsymbol{w}$,
drawn from $q(\boldsymbol{w};\boldsymbol{\phi})$. This is \emph{equivalent
to minimizing a KL divergence measure} between the inferred approximate
variational density and the actual underlying distribution.

Turning to the selection of the imposed prior $p(\boldsymbol{w})$,
one may opt for a fixed-form \emph{isotropic }Gaussian:
\begin{equation}
p(\boldsymbol{w})=\mathcal{N}(\boldsymbol{w}|0,\sigma_{0}^{2}\boldsymbol{I})
\end{equation}
where $\boldsymbol{I}$ is the identity matrix, and $\mathcal{N}(\cdot|\boldsymbol{\mu},\boldsymbol{\Sigma})$
is a multivariate Gaussian with mean $\boldsymbol{\mu}$ and covariance
matrix $\boldsymbol{\Sigma}$. On the other hand, the sought variational
posterior $q(\boldsymbol{w};\boldsymbol{\phi})$ is for simplicity
and efficiency purposes selected as a diagonal Gaussian of the form:
\begin{equation}
q(\boldsymbol{w};\boldsymbol{\phi})=\mathcal{N}(\boldsymbol{w}|\boldsymbol{\mu},\mathrm{diag}(\boldsymbol{\sigma}^{2}))
\end{equation}
 where $\boldsymbol{\phi}=\{\boldsymbol{\mu},\boldsymbol{\sigma}^{2}\}$,
and $\mathrm{diag}(\boldsymbol{\sigma}^{2})$ is a diagonal matrix
with the vector $\boldsymbol{\sigma}^{2}$ on its main diagonal.

An issue with the above formulation is that the entailed posterior
expectation $\mathbb{E}_{q(\boldsymbol{w};\boldsymbol{\phi})}[\mathrm{log}\,p(\mathcal{D}|\boldsymbol{w})]$
is analytically intractable; this is due to the non-conjugate nature
of deep networks, stemming from the employed nonlinear activation
functions. This prohibits taking derivatives of $\mathcal{L}(\boldsymbol{\phi})$
to effect derivation of the sought posterior $q(\boldsymbol{w};\boldsymbol{\phi})$.
In addition, approximating this expectation by simply drawing Monte-Carlo
(MC) samples from the weights posterior is not an option, due to the
prohibitively high variance of the resulting estimator. 

To address this issue, one can resort to a simple reparameterization
trick: We consider that the MC samples $\boldsymbol{w}^{(s)}$ used
to approximate the expectation $\mathbb{E}_{q(\boldsymbol{w};\boldsymbol{\phi})}[\mathrm{log}\,p(\mathcal{D}|\boldsymbol{w})]$
are functions of their posterior mean and variance, as well as a random
noise vector, $\boldsymbol{\epsilon}$, sampled from a standard Gaussian
distribution \cite{aevb4,aevb3,aevb}. This can be effected by introducing
the transform:
\begin{equation}
\boldsymbol{w}=\boldsymbol{\mu}+\boldsymbol{\sigma}\varodot\boldsymbol{\epsilon}
\end{equation}
where $\varodot$ denotes the elementwise product of two vectors,
and the $\boldsymbol{\epsilon}$ are distributed as $\boldsymbol{\epsilon}\sim\mathcal{N}(\boldsymbol{0},\boldsymbol{I})$.
By substituting this transform into the derived ELBO expression, the
entailed posterior expectation is expressed as an average over a standard
Gaussian density, $p(\boldsymbol{\epsilon})$. This yields an MC estimator
with low variance, under some mild conditions \cite{aevb}.

\subsection{The Student's-$t$ Distribution}

The adoption of the multivariate Student's-$t$ distribution provides
a way to broaden the Gaussian distribution for potential outliers
\cite[Section 7]{key-19}. The probability density function (pdf)
of a Student's-$t$ distribution with mean vector $\boldsymbol{\mu}$,
covariance matrix $\boldsymbol{\Sigma}$, and $\nu>0$ degrees of
freedom is \cite{liu}
\begin{equation}
t(\boldsymbol{y}_{t};\boldsymbol{\mu},\boldsymbol{\Sigma},\nu)=\frac{\Gamma\left(\frac{\nu+\delta}{2}\right)|\boldsymbol{\Sigma}|^{-1/2}(\pi\nu)^{-\delta/2}}{\Gamma(\nu/2)\{1+d(\boldsymbol{y}_{t},\boldsymbol{\mu};\boldsymbol{\Sigma})/\nu\}^{(\nu+\delta)/2}}
\end{equation}
where $\delta$ is the dimensionality of the observations $\boldsymbol{y}_{t},$
$d(\boldsymbol{y}_{t},\boldsymbol{\mu};\boldsymbol{\Sigma})$ is the
squared Mahalanobis distance between $\boldsymbol{y}_{t},\boldsymbol{\mu}$
with covariance matrix $\boldsymbol{\Sigma}$
\begin{equation}
d(\boldsymbol{y}_{t},\boldsymbol{\mu};\boldsymbol{\Sigma})=(\boldsymbol{y}_{t}-\boldsymbol{\mu})^{T}\boldsymbol{\Sigma}^{-1}(\boldsymbol{y}_{t}-\boldsymbol{\mu})
\end{equation}
and $\Gamma(s)$ is the Gamma function, $\Gamma(s)=\int_{0}^{\infty}e^{-t}z^{s-1}dz$. 

It can be shown (see, e.g., \cite{liu}) that, in essence, the Student's-$t$
distribution corresponds to a Gaussian scale model \cite{gsm} where
the precision scalar is a Gamma distributed latent variable, depending
on the degrees of freedom of the Student's-$t$ density. That is,
given
\begin{equation}
\boldsymbol{y}_{t}\sim t(\boldsymbol{\mu},\boldsymbol{\Sigma},\nu)
\end{equation}
 it equivalently holds that \cite{liu}
\begin{equation}
\boldsymbol{y}_{t}|\xi_{t}\sim\mathcal{N}(\boldsymbol{\mu},\boldsymbol{\Sigma}/\xi_{t})
\end{equation}
 where the precision scalar, $\xi_{t}$, is distributed as
\begin{equation}
\xi_{t}\sim\mathcal{G}\left(\frac{\nu}{2},\frac{\nu}{2}\right)
\end{equation}
and $\mathcal{G}(\alpha,\beta)$ is the Gamma distribution. 

A graphical illustration of the univariate Student's-$t$ distribution,
with $\boldsymbol{\mu}$, $\boldsymbol{\Sigma}$ fixed, and for various
values of the degrees of freedom $\nu$, is provided in Fig. 2. As
we observe, as $\nu\rightarrow\infty$, the Student's-$t$ distribution
tends to a Gaussian with the same $\boldsymbol{\mu}$ and $\boldsymbol{\Sigma}$.
On the contrary, as $\nu$ tends to zero, the tails of the distribution
become longer, thus allowing for a better handling of potential outliers,
without affecting the mean or the covariance of the distribution.
Thus, by exploiting the heavier tails of the Student's-$t$ distribution,
a probabilistic generative model becomes capable of handling, in a
considerably enhanced manner, outliers residing in the fitting datasets.
That is, if the modeled phenomenon is actually heavy-tailed, the inferred
Student's-$t$ model will be capable to cope, by yielding a value
for the fitted degrees of freedom parameter, $\nu$, e.g., close to
1. On the other hand, if no such heavy-tailed nature does actually
characterize the data, the fitted degrees of freedom parameter, $\nu$,
will yield a value close to infinity (practically, above 100). In
the latter case, the model essentially reduces to a simpler Gaussian
density \cite[Section 7]{key-19}.

\begin{figure}
\begin{centering}
\includegraphics[scale=0.45]{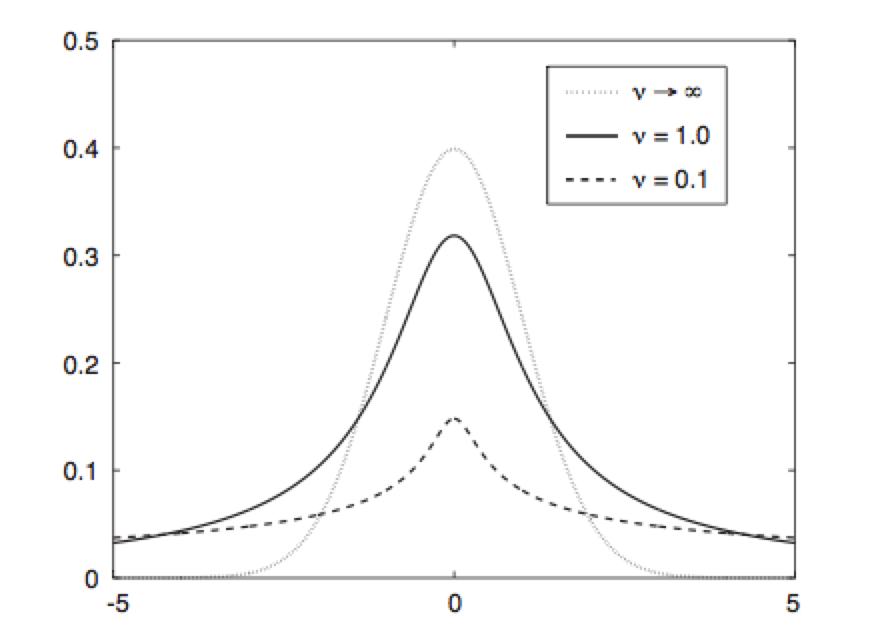} 
\par\end{centering}
\caption{Univariate Student's-$t$ distribution $t(\boldsymbol{y}_{t};\boldsymbol{\mu},\boldsymbol{\Sigma},\nu)$,
with $\boldsymbol{\mu}$, $\boldsymbol{\Sigma}$ fixed, for various
values of $\nu$ \cite{vbsmm2}. }
\end{figure}

\subsection{The $t$-Divergence}

As discussed in Section II.B, conventional variational inference is
equivalent to minimization of a KL divergence measure, which is also
known as the relative SBG-entropy. Motivated from these facts, and
in order to allow for making the most out of the merits (heavy tails)
of the $t$-exponential family, the $t$-divergence was introduced
in \cite{tVB} as follows:\\
\textbf{Definition 1. }The $t$-divergence between two distributions,
$q(\boldsymbol{h})$ and $p(\boldsymbol{h})$, is defined as
\begin{equation}
D_{t}(q||p)=\int\tilde{q}(\boldsymbol{h})\mathrm{log}_{t}q(\boldsymbol{h})\mathrm{d}\boldsymbol{h}-\tilde{q}(\boldsymbol{h})\mathrm{log}_{t}p(\boldsymbol{h})\mathrm{d}\boldsymbol{h}
\end{equation}
 where $\tilde{q}(\boldsymbol{h})$ is called the escort distribution
of $q(\boldsymbol{h})$, and is given by
\begin{equation}
\tilde{q}(\boldsymbol{h})=\frac{q(\boldsymbol{h})^{t}}{\int q(\boldsymbol{h})^{t}\mathrm{d}\boldsymbol{h}},\;t\in\mathbb{R}
\end{equation}
Importantly, the divergence $D_{t}(q||p)$ preserves the following
two properties:
\begin{itemize}
\item $D_{t}(q||p)\geq0,\;\forall q,p$. The equality holds only for $q=p$.
\item $D_{t}(q||p)\neq D_{t}(p||q)$. 
\end{itemize}
In the seminal work of \cite{tVB}, it has been shown that by leveraging
the above definition of the $t$-divergence, $D_{t}(q||p)$, one can
establish an advanced variational inference framework, much more appropriate
for modeling data with heavy tails. We exploit these benefits in developing
the training and inference algorithms of the proposed $t$-MEM-NN
model, as explained in the following Section. 

\section{Proposed Approach}

\subsection{Model Formulation}

$t$-MEM-NN extends upon the model design principles discussed previously,
by building on the solid theory of variational inference based on
the $t$-divergence. It does so by introducing a novel formulation
that renders MEM-NN amenable to Bayesian inference. 

To effect our modeling goals, we first consider that the postulated
embeddings matrices are Student's-$t$ distributed latent variables.
Specifically, let us start by imposing a simple, zero-mean Student's-$t$
prior distribution over them, with tied degrees of freedom:
\begin{equation}
p(\boldsymbol{A})=t(\mathrm{vec}(\boldsymbol{A})|\boldsymbol{0},\boldsymbol{I},\nu)
\end{equation}
\begin{equation}
p(\boldsymbol{B})=t(\mathrm{vec}(\boldsymbol{B})|\boldsymbol{0},\boldsymbol{I},\nu)
\end{equation}
\begin{equation}
p(\boldsymbol{C})=t(\mathrm{vec}(\boldsymbol{C})|\boldsymbol{0},\boldsymbol{I},\nu)
\end{equation}
where $\mathrm{vec}(\cdot)$ is the matrix vectorization operation,
and $\nu>0$ is the degrees of freedom hyperparameter of the imposed
priors. On this basis, we seek to devise an efficient means of inferring
the corresponding posterior distributions, given the available training
data. We postulate that the sought posterior $q(\boldsymbol{A},\boldsymbol{B},\boldsymbol{C};\boldsymbol{\phi})$
factorizes over $\boldsymbol{A}$, $\boldsymbol{B}$, and $\boldsymbol{C}$
(mean-field approximation \cite{naudts2}); the factors are considered
to approximately take the form of Student's-$t$ densities with means,
\emph{diagonal} covariance matrices, and degrees of freedom inferred
from the data. Hence, we have: 
\begin{equation}
q(\boldsymbol{A};\boldsymbol{\phi})=t(\mathrm{vec}(\boldsymbol{A})|\boldsymbol{\mu}_{\boldsymbol{A}},\mathrm{diag}(\boldsymbol{\sigma}_{\boldsymbol{A}}^{2}),\nu_{\boldsymbol{A}})
\end{equation}

\begin{equation}
q(\boldsymbol{B};\boldsymbol{\phi})=t(\mathrm{vec}(\boldsymbol{B})|\boldsymbol{\mu}_{\boldsymbol{B}},\mathrm{diag}(\boldsymbol{\sigma}_{\boldsymbol{B}}^{2}),\nu_{\boldsymbol{B}})
\end{equation}

\begin{equation}
q(\boldsymbol{C};\boldsymbol{\phi})=t(\mathrm{vec}(\boldsymbol{C})|\boldsymbol{\mu}_{\boldsymbol{C}},\mathrm{diag}(\boldsymbol{\sigma}_{\boldsymbol{C}}^{2}),\nu_{\boldsymbol{C}})
\end{equation}
 where $\boldsymbol{\phi}=\{\boldsymbol{\mu}_{i},\boldsymbol{\sigma}_{i}^{2},\nu_{i}\}_{i\in\{\boldsymbol{A},\boldsymbol{B},\boldsymbol{C}\}}$,
and $\nu_{i}>0,\forall i$.

On this basis, to perform model training in a way the best exploits
the heavy tails of the developed model, we minimize the $t$-divergence
between the sought variational posterior and the postulated joint
density over the observed data and the model latent variables. Thus,
the proposed model training objective becomes
\begin{equation}
\small\begin{aligned}q(\boldsymbol{A} & ;\boldsymbol{\phi}),q(\boldsymbol{B};\boldsymbol{\phi}),q(\boldsymbol{C};\boldsymbol{\phi}),\boldsymbol{W}\\
= & \underset{q,\boldsymbol{W}}{\mathrm{arg\,min}\;}D_{t}\left(q(\boldsymbol{A};\boldsymbol{\phi}),q(\boldsymbol{B};\boldsymbol{\phi}),q(\boldsymbol{C};\boldsymbol{\phi})||p(a;\boldsymbol{A},\boldsymbol{B},\boldsymbol{C},\boldsymbol{W})\right)
\end{aligned}
\end{equation}
where $p(a;\boldsymbol{A},\boldsymbol{B},\boldsymbol{C},\boldsymbol{W})=p(a;\boldsymbol{W})p(\boldsymbol{A})p(\boldsymbol{B})p(\boldsymbol{C})$.
Then, following the derivations rationale of \cite{tVB}, and by application
of  simple algebra, the expression of the $t$-divergence in (22)
yields
\begin{equation}
\begin{aligned}D_{t}\big(q(\boldsymbol{A};\boldsymbol{\phi}),q(\boldsymbol{B};\boldsymbol{\phi}),q(\boldsymbol{C};\boldsymbol{\phi}) & ||p(a;\boldsymbol{A},\boldsymbol{B},\boldsymbol{C},\boldsymbol{W})\big)=\\
=D_{t}\left(q(\boldsymbol{A};\boldsymbol{\phi})||p(\boldsymbol{A})\right) & +D_{t}\left(q(\boldsymbol{B};\boldsymbol{\phi})||p(\boldsymbol{B})\right)\\
+D_{t}\left(q(\boldsymbol{C};\boldsymbol{\phi})||p(\boldsymbol{C})\right) & -\mathbb{E}_{\tilde{q}(\cdot;\boldsymbol{\phi})}[\mathrm{log}p(a;\boldsymbol{W})]
\end{aligned}
\end{equation}
 where $\tilde{q}(\cdot;\boldsymbol{\phi})$ is the escort distribution
of the sought variational posterior, and $p(a;\boldsymbol{W})$ is
a Multinoulli parameterized via the probability vector $\hat{\boldsymbol{a}}$,
given by (4).

Following \cite{tVB}, and based on (16)-(21), we obtain that the
$t$-divergence expressions in (23) can be written in the following
form:
\begin{equation}
\begin{aligned}D_{t}\left(q(\boldsymbol{\Theta};\boldsymbol{\phi})||p(\boldsymbol{\Theta})\right)=\sum_{l=1}^{\delta V}\bigg\{ & \frac{\Psi_{ql}}{1-t}\left(1+\frac{1}{\nu_{\boldsymbol{\Theta}}}\right)\\
-\frac{\Psi_{p}}{1-t} & \left(1+\frac{[\boldsymbol{\sigma}_{\boldsymbol{\Theta}}^{2}]_{l}+[\boldsymbol{\mu_{\Theta}}]_{l}^{2}}{\nu}\right)\bigg\}
\end{aligned}
\end{equation}
where $\boldsymbol{\Theta}\in\{\boldsymbol{A},\boldsymbol{B},\boldsymbol{C}\}$,
$[\boldsymbol{\xi}]_{l}$ is the $l$th element of a vector $\boldsymbol{\xi}$,
we denote
\begin{equation}
\Psi_{ql}\triangleq\left(\frac{\mathrm{\Gamma}(\frac{\nu_{\boldsymbol{\Theta}}+1}{2})}{\mathrm{\Gamma}(\frac{\nu_{\boldsymbol{\Theta}}}{2})(\pi\nu_{\boldsymbol{\Theta}})^{1/2}[\boldsymbol{\sigma}_{\boldsymbol{\Theta}}]_{l}}\right)^{-\frac{2}{\nu_{\boldsymbol{\Theta}}+1}}
\end{equation}
\begin{equation}
\Psi_{p}\triangleq\left(\frac{\mathrm{\Gamma}(\frac{\nu+1}{2})}{\mathrm{\Gamma}(\frac{\nu}{2})(\pi\nu)^{1/2}}\right)^{-\frac{2}{\nu+1}}
\end{equation}
 $\delta$ is the dimensionality of the embeddings, $V$ is the vocabulary
size, $\nu$ is the degrees of freedom hyperparameter of the prior,
and the free hyperparameter $t$ is set as \cite{tVB}
\begin{equation}
t=\frac{2}{1+\nu_{\boldsymbol{\Theta}}}+1
\end{equation}

\subsection{Training Algorithm Configuration}

As we observe from the preceding discussion, the expectation of the
conditional log-likelihood of the model, $\mathbb{E}_{\tilde{q}(\cdot;\boldsymbol{\phi})}[\mathrm{log}p(a;\boldsymbol{W})]$,
is computed with respect to the escort distribution of the sought
posterior, $\tilde{q}(\cdot;\boldsymbol{\phi})$. Based on (19)-(21),
it is easy to show that this escort distribution yields a factorized
form, with \cite{tVB}
\begin{equation}
\begin{aligned}\tilde{q}(\boldsymbol{\Theta};\boldsymbol{\phi})=t\left(\mathrm{vec}(\boldsymbol{\Theta})|\boldsymbol{\mu_{\Theta}},\frac{\nu_{\boldsymbol{\Theta}}}{\nu_{\boldsymbol{\Theta}}+2}\mathrm{diag}(\boldsymbol{\sigma}_{\boldsymbol{\Theta}}^{2}),\nu_{\boldsymbol{\Theta}}+2\right)\\
\forall\boldsymbol{\Theta}\in\{\boldsymbol{A},\boldsymbol{B},\boldsymbol{C}\}
\end{aligned}
\end{equation}
Despite this convenient escort distribution expression, though, this
posterior expectation cannot be computed analytically; hence, its
gradient becomes intractable. This is due to the nonconjugate nature
of $t$-MEM-NN, which stems from its nonlinear assumptions. Apparently,
approximating this expectation using a set of $S$ MC samples, $\{\boldsymbol{\Theta}_{s}\}_{s=1}^{S}$,
drawn from the escort densities (28), would result in estimators with
unacceptably high variance. 

In this work, these issues are resolved by adopting the reparameterization
trick ideas described in Section II.B, adapted to the $t$-exponential
family. Specifically, we perform a smart reparameterization of the
MC samples from the Student's-$t$ escort densities (28) which yields:
\begin{equation}
\boldsymbol{\Theta}_{s}=\boldsymbol{\mu}_{\boldsymbol{\Theta}}+\left(\frac{\nu_{\boldsymbol{\Theta}}}{\nu_{\boldsymbol{\Theta}}+2}\right)^{1/2}\boldsymbol{\sigma}_{\boldsymbol{\Theta}}\boldsymbol{\epsilon}_{s}
\end{equation}
where $\boldsymbol{\epsilon}_{s}$ is random Student's-$t$ noise
with unitary variance:
\begin{equation}
\boldsymbol{\epsilon}_{s}\sim t(\boldsymbol{0},\boldsymbol{I},\nu_{\boldsymbol{\Theta}}+2)
\end{equation}

Then, the resulting (reparameterized) $t$-divergence objective (23)
can be minimized by means of any off-the-shelf stochastic optimization
algorithm. For this purpose, in this work we utilize Adagrad; this
constitutes a stochastic gradient descent algorithm with adaptive
step-size \cite{adagrad}, and fast and proven convergence to a local
optimum. Adagrad updates the trained posterior hyperparameter set,
$\boldsymbol{\phi}$, as well as the output layer weights, $\boldsymbol{W}$,
by utilizing the gradient $\triangledown_{\boldsymbol{\phi},\boldsymbol{W}}D_{t}\left(q(\boldsymbol{A};\boldsymbol{\phi}),q(\boldsymbol{B};\boldsymbol{\phi}),q(\boldsymbol{C};\boldsymbol{\phi})||p(a;\boldsymbol{A},\boldsymbol{B},\boldsymbol{C},\boldsymbol{W})\right)$. 

On each Adagrad iteration, this gradient is computed using only a
small subset (minibatch) of the available training data, as opposed
to using the whole training dataset. This allows for computational
tractability, no matter what the total number of training examples
is. To facilitate convergence, on each algorithm iteration a different
minibatch is selected, in a completely random fashion. 

In this context, it is important to appropriately select the number
of MC samples drawn from (30) \emph{during training}. In our work,
we opt for the computationally efficient solution of drawing \emph{just
one MC sample} \emph{during training}. One could argue that using
only one MC sample is doomed to result in an approximation of limited
quality. However, it has been empirically well-established that drawing
just one MC sample is sufficient when Adagrad is executed with a small
minibatch size compared to the size of the used training dataset \cite{vb-weights,asydgm}.
Indeed, this is the case with our experimental evaluations in Section
IV. In all cases, network initialization is performed by means of
the Glorot uniform scheme, except for the degrees of freedom hyperparameters;
these are initialized at high values ($\nu=100$), which essentially
reduce the initial Student's-$t$ densities of our model to simpler
Gaussian ones (as discussed in Section II.C) \cite{glorot}. 

\subsection{Inference Procedure}

Having obtained a training algorithm for our proposed $t$-MEM-NN
model, we can now proceed to elaborate on how inference is performed
using our method. As briefly hinted in Section II.B, this consists
in drawing a number of MC samples from the inferred posteriors over
the model parameters, $q(\cdot;\boldsymbol{\phi}),$ and obtaining
the average predictive value of the model that corresponds to these
drawn parameter values (samples). According to the related deep learning
literature, drawing a set of $S=10$ samples should be enough \emph{for
inference purposes} \cite{vb-weights,asydgm}. We investigate the
impact of the number of drawn MC samples to the eventually obtained
performance of the inference algorithm of our model in our experiments
that follow.

\section{Synthetic Question-Answering Tasks}

\begin{table*}
\centering{}\caption{Considered benchmark dataset: Indicative training data samples from
three of the entailed types of tasks.}
\begin{tabular}{l|l|l}
{\small{}{}Sam walks into the kitchen.}  & {\small{}{}Brian is a lion.}  & {\small{}{}Sandra got the milk.}\tabularnewline
{\small{}{}Sam picks up an apple.}  & {\small{}{}Julius is a lion.}  & {\small{}{}Sandra journeyed to the garden.}\tabularnewline
{\small{}{}Sam walks into the bedroom.}  & {\small{}{}Julius is white.}  & {\small{}{}Sandra went back to the bathroom.}\tabularnewline
{\small{}{}Sam drops the apple.}  & {\small{}{}Bernhard is green.}  & {\small{}{}Sandra put down the milk.}\tabularnewline
\textcolor{blue}{\small{}{}Q: Where is the apple?}{\small{} } & \textcolor{blue}{\small{}{}Q: What color is Brian?}{\small{} } & \textcolor{blue}{\small{}{}Q: Where was the milk before the bathroom?}\tabularnewline
\textcolor{red}{\small{}{}A: Bedroom}{\small{} } & \textcolor{red}{\small{}{}A: White}{\small{} } & \textcolor{red}{\small{}{}A: Garden}\tabularnewline
\end{tabular}
\end{table*}

\begin{table}
\caption{Quantitative assessment: Accuracy results in the test set.}

\centering{}%
\begin{tabular}{|l||c||c||c|}
\hline 
\multicolumn{4}{|c|}{Test Accuracy (\%)}\tabularnewline
\hline 
\hline 
 & Baseline & \multicolumn{2}{c|}{$t$-MEM-NN}\tabularnewline
\hline 
Task type & MemN2N & 1 sample & 10 samples\tabularnewline
\hline 
1: 1 supporting fact  & 100 & 100 & 100\tabularnewline
2: 2 supporting facts  & 84 & 77 & 84\tabularnewline
3: 3 supporting facts & 55 & 55 & \textbf{57}\tabularnewline
4: 2 argument relations & 96 & 94 & 96\tabularnewline
5: 3 argument relations & 88 & 87 & 88\tabularnewline
6: yes/no questions & 92 & 93 & \textbf{97}\tabularnewline
7: counting & 83 & 81 & \textbf{85}\tabularnewline
8: lists/sets & 87 & 87 & \textbf{90}\tabularnewline
9: simple negation & 90 & 88 & \textbf{92}\tabularnewline
10: indefinite knowledge & 78 & 81 & \textbf{84}\tabularnewline
11: basic coreference & 85 & 98 & \textbf{98}\tabularnewline
12: conjunction & 100 & 100 & 100\tabularnewline
13: compound coreference & 89 & 93 & \textbf{95}\tabularnewline
14: time reasoning & 92 & 92 & \textbf{96}\tabularnewline
15: basic deduction & 100 & 100 & 100\tabularnewline
16: basic induction & 45 & 45 & \textbf{46}\tabularnewline
17: positional reasoning & 51 & 51  & \textbf{53}\tabularnewline
18: size reasoning & 87 & 89 & \textbf{91}\tabularnewline
19: path finding & 12 & 12  & \textbf{14}\tabularnewline
20: agent’s motivation & 100  & 100 & 100\tabularnewline
\hline 
\end{tabular}
\end{table}

\begin{figure}
\begin{centering}
\includegraphics[scale=0.5]{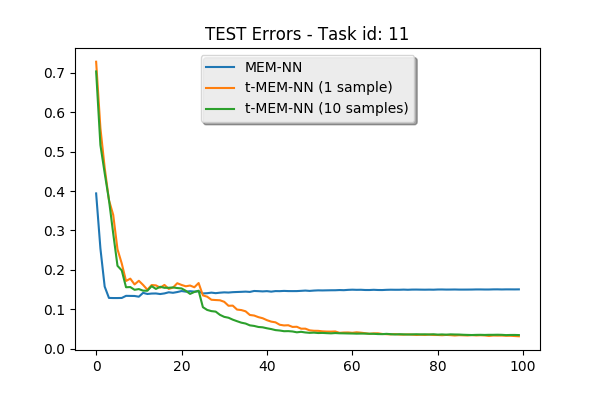}
\par\end{centering}
\caption{Test error per epoch: Task type \#11.}
\end{figure}

\begin{table*}
\caption{Attention in task type \#1 - story \#202.}

\begin{centering}
\begin{tabular}{|l|}
\hline 
\multicolumn{1}{|c|}{Facts}\tabularnewline
\hline 
\hline 
1. mary moved to the hallway \tabularnewline
1. daniel travelled to the office \tabularnewline
2. john went back to the hallway \tabularnewline
3. john moved to the office\tabularnewline
4. sandra journeyed to the kitchen\tabularnewline
5. mary moved to the bedroom\tabularnewline
\hline 
\end{tabular}%
\begin{tabular}{c}
Question\tabularnewline
\hline 
where is daniel?\tabularnewline
\end{tabular}%
\begin{tabular}{c}
Answer\tabularnewline
\hline 
office\tabularnewline
\end{tabular}%
\begin{tabular}{c}
Supporting Facts\tabularnewline
\hline 
daniel travelled to the office\tabularnewline
\end{tabular}
\par\end{centering}
\begin{centering}
\subfloat[Predicted answers.]{

\begin{tabular}{|c|}
\hline 
\textit{\emph{MemN2N}}\tabularnewline
\hline 
\hline 
\textcolor{red}{bedroom}\tabularnewline
\hline 
\end{tabular}%
\begin{tabular}{|c|}
\hline 
$t$-MEM-NN (1 sample)\tabularnewline
\hline 
\hline 
\textcolor{red}{bedroom}\tabularnewline
\hline 
\end{tabular}%
\begin{tabular}{|c|}
\hline 
$t$-MEM-NN (10 samples)\tabularnewline
\hline 
\hline 
\textcolor{green}{office}\tabularnewline
\hline 
\end{tabular}}
\par\end{centering}
\centering{}\subfloat[Model attention per hop.]{

\begin{tabular}{|c||c|}
\hline 
\multicolumn{2}{|c|}{\textit{\emph{MemN2N}}}\tabularnewline
\hline 
\hline 
hop 1 & \textcolor{green}{daniel travelled to the office}\tabularnewline
\hline 
hop 2 & \textcolor{green}{daniel travelled to the office}\tabularnewline
\hline 
hop 3 & \textcolor{red}{mary moved to the bedroom}\tabularnewline
\hline 
\end{tabular}%
\begin{tabular}{|c||c|}
\hline 
\multicolumn{2}{|c|}{$t$-MEM-NN (1 sample)}\tabularnewline
\hline 
\hline 
hop 1 & \textcolor{green}{daniel travelled to the office}\tabularnewline
\hline 
hop 2 & \textcolor{green}{daniel travelled to the office}\tabularnewline
\hline 
hop 3 & \textcolor{red}{mary moved to the bedroom}\tabularnewline
\hline 
\end{tabular}%
\begin{tabular}{|c||c|}
\hline 
\multicolumn{2}{|c|}{$t$-MEM-NN (10 samples)}\tabularnewline
\hline 
\hline 
hop 1 & \textcolor{green}{daniel travelled to the office}\tabularnewline
\hline 
hop 2 & \textcolor{green}{daniel travelled to the office}\tabularnewline
\hline 
hop 3 & \textcolor{green}{daniel travelled to the office}\tabularnewline
\hline 
\end{tabular}}
\end{table*}

\begin{table*}
\begin{centering}
\caption{Attention in task type \#11 - story \#3.}
\begin{tabular}{|l|}
\hline 
\multicolumn{1}{|c|}{Facts}\tabularnewline
\hline 
\hline 
1. john journeyed to the hallway \tabularnewline
2. after that he journeyed to the garden\tabularnewline
3. john moved to the office \tabularnewline
4. following that he went to the hallway\tabularnewline
5. sandra travelled to the bedroom\tabularnewline
6. then she moved to the hallway \tabularnewline
7. mary travelled to the hallway\tabularnewline
8. afterwards she went to the bathroom\tabularnewline
\hline 
\end{tabular}%
\begin{tabular}{c}
Question\tabularnewline
\hline 
where is sandra?\tabularnewline
\tabularnewline
\end{tabular}%
\begin{tabular}{c}
Answer\tabularnewline
\hline 
hallway\tabularnewline
\tabularnewline
\end{tabular}%
\begin{tabular}{c}
Supporting Facts\tabularnewline
\hline 
sandra travelled to the bedroom\tabularnewline
then she moved to the hallway\tabularnewline
\end{tabular}
\par\end{centering}
\begin{centering}
\subfloat[Predicted answers.]{

\begin{tabular}{|c|}
\hline 
\textit{\emph{MemN2N}}\tabularnewline
\hline 
\textcolor{red}{bathroom}\tabularnewline
\hline 
\end{tabular}%
\begin{tabular}{|c|}
\hline 
$t$-MEM-NN (1 sample)\tabularnewline
\hline 
\textcolor{green}{hallway}\tabularnewline
\hline 
\end{tabular}%
\begin{tabular}{|c|}
\hline 
$t$-MEM-NN (10 samples)\tabularnewline
\hline 
\textcolor{green}{hallway}\tabularnewline
\hline 
\end{tabular}}
\par\end{centering}
\centering{}\subfloat[Model attention per hop.]{

\begin{tabular}{|c||l|}
\hline 
\multicolumn{2}{|c|}{\textit{\emph{MemN2N}}}\tabularnewline
\hline 
\hline 
hop 1 & \textcolor{red}{afterwards she went to the bathroom}\tabularnewline
\hline 
hop 2 & \textcolor{red}{afterwards she went to the bathroom}\tabularnewline
\hline 
hop 3 & \textcolor{red}{afterwards she went to the bathroom}\tabularnewline
\hline 
\end{tabular}%
\begin{tabular}{|c||l|}
\hline 
\multicolumn{2}{|c|}{$t$-MEM-NN (1 sample)}\tabularnewline
\hline 
\hline 
hop 1 & \textcolor{green}{sandra travelled to the bedroom}\tabularnewline
\hline 
hop 2 & \textcolor{green}{then she moved to the hallway}\tabularnewline
\hline 
hop 3 & \textcolor{green}{then she moved to the hallway}\tabularnewline
\hline 
\end{tabular}%
\begin{tabular}{|c||l|}
\hline 
\multicolumn{2}{|c|}{$t$-MEM-NN (10 samples)}\tabularnewline
\hline 
\hline 
hop 1 & \textcolor{green}{sandra travelled to the bedroom}\tabularnewline
\hline 
hop 2 & \textcolor{green}{then she moved to the hallway}\tabularnewline
\hline 
hop 3 & \textcolor{green}{then she moved to the hallway}\tabularnewline
\hline 
\end{tabular}}
\end{table*}

\begin{table*}
\caption{Attention in task type \#14 - story \#22.}
\begin{tabular}{|l|}
\hline 
\multicolumn{1}{|c|}{Facts}\tabularnewline
\hline 
\hline 
1. mary went back to the kitchen this morning \tabularnewline
2. mary travelled to the school yesterday\tabularnewline
3. yesterday fred travelled to the bedroom\tabularnewline
4. yesterday bill moved to the park \tabularnewline
5. this afternoon bill went back to the park\tabularnewline
6. bill went to the school this morning\tabularnewline
\hline 
\end{tabular}%
\begin{tabular}{c}
Question\tabularnewline
\hline 
where was bill before the park?\tabularnewline
\tabularnewline
\end{tabular}%
\begin{tabular}{c}
Answer\tabularnewline
\hline 
school\tabularnewline
\tabularnewline
\end{tabular}%
\begin{tabular}{c}
Supporting Facts\tabularnewline
\hline 
this afternoon bill went back to the park\tabularnewline
bill went to the school this morning\tabularnewline
\end{tabular}
\begin{centering}
\subfloat[Predicted answers.]{

\begin{tabular}{|c|}
\hline 
\textit{\emph{MemN2N}}\tabularnewline
\hline 
\textcolor{red}{park}\tabularnewline
\hline 
\end{tabular}%
\begin{tabular}{|c|}
\hline 
$t$-MEM-NN (1 sample)\tabularnewline
\hline 
\textcolor{red}{park}\tabularnewline
\hline 
\end{tabular}%
\begin{tabular}{|c|}
\hline 
$t$-MEM-NN (10 samples)\tabularnewline
\hline 
\textcolor{green}{school}\tabularnewline
\hline 
\end{tabular}}
\par\end{centering}
\centering{}\subfloat[Model attention per hop.]{

\begin{tabular}{|c||l|}
\hline 
\multicolumn{2}{|c|}{\textit{\emph{MemN2N}}}\tabularnewline
\hline 
\hline 
hop 1 & \textcolor{red}{yesterday bill moved to the park}\tabularnewline
\hline 
hop 2 & \textcolor{red}{yesterday bill moved to the park}\tabularnewline
\hline 
hop 3 & \textcolor{red}{yesterday bill moved to the park}\tabularnewline
\hline 
\end{tabular}%
\begin{tabular}{|c||l|}
\hline 
\multicolumn{2}{|c|}{$t$-MEM-NN (1 sample)}\tabularnewline
\hline 
\hline 
hop 1 & \textcolor{red}{yesterday bill moved to the park}\tabularnewline
\hline 
hop 2 & \textcolor{green}{this afternoon bill went back to the park}\tabularnewline
\hline 
hop 3 & \textcolor{red}{yesterday bill moved to the park}\tabularnewline
\hline 
\end{tabular}%
\begin{tabular}{|c||l|}
\hline 
\multicolumn{2}{|c|}{$t$-MEM-NN (10 samples)}\tabularnewline
\hline 
\hline 
hop 1 & \textcolor{red}{yesterday bill moved to the park}\tabularnewline
\hline 
hop 2 & \textcolor{green}{this afternoon bill went back to the park}\tabularnewline
\hline 
hop 3 & \textcolor{green}{bill went to the school this morning}\tabularnewline
\hline 
\end{tabular}}
\end{table*}

In this Section, we perform a thorough experimental evaluation of
our proposed $t$-MEM-NN model. We provide a quantitative assessment
of the efficacy, the effectiveness, and the computational efficiency
of our approach, combined with deep qualitative insights into few
of its key performance characteristics. To this end, we utilize a
publicly available benchmark, which is popular in the recent literature,
namely the set of synthetic question-answering (QA) tasks defined
in \cite{babi} (bAbI). Specifically, we consider the \emph{English-Language
}tasks of the bAbI dataset that comprise 1K training examples (\emph{en-1K}
tasks). This dataset comprises 20 different types of tasks, which
are characterized by different qualitative properties. Some of them
are harder to be learned, while some others are much easier. 

The idea of all the types of tasks entailed in this dataset is rather
simple. Each task consists of a set of statements (facts), a question,
and an answer. The answer comprises only one word from the available
vocabulary. Given the facts, a question is asked and an answer is
expected. Then, model performance can be evaluated on the basis of
the percentage of generated answers that match the available groundtruth.
To allow for a better feeling of what our dataset looks like, we provide
indicative samples from three of the entailed types of tasks in Table
I. Note that the available dataset also provides additional supporting
facts that may be made use of by the trained models. 

To provide some comparative results, we evaluate two variants of our
method, namely one where \emph{inference is performed} using only
a single MC sample (drawn from the model posteriors), and another
one where 10 MC samples are used. In all cases, \emph{training }is
performed by drawing just one MC sample. In addition, we compare to
the state-of-the-art alternative that is the closest related to our
approach, namely the MemN2N method of \cite{n2n-mem}. Our source
codes have been developed in Python, using the TensorFlow library
\cite{tensorflow2015-whitepaper}, as well as open-source software
published by Dominique Luna\footnote{\url{https://github.com/domluna/memn2n}}.
Our experiments are run on an Intel Xeon server with 64GB RAM and
an NVIDIA Tesla K40 GPU.

\subsection{Experimental Setup}

Model training is performed by utilizing the training dataset provided
in the used \emph{en-1K} bAbI benchmark. This comprises 1000 examples
from each type of task; from these, we randomly select 900 samples
to perform training, and retain the remainder 100 for validation purposes.
Each training example comprises the full set of data pertaining to
the task, including the correct answer (which we expect the system
to generate), apart from the corresponding question and available
statements (facts). On this basis, a trained model is evaluated by
presenting it with the facts and the questions pertaining to each
example in the test set, and running its inference algorithm to obtain
a predicted answer. The available test set comprises 1000 cases from
each task type, which are completely unknown to the trained models\emph{. }

Since the used benchmark comprises a multitude of task types, we train
a different model (of each evaluated method) for each task type\emph{.
}An obvious advantage of such a modeling setup consists in the fact
that it allows for the trained models to be finely-tuned to data with
very specific patterns. On the other hand, the imposed weight tying
across model layers is a strong safeguard against possible model overfitting. 

Turning to the selection of the hyperparameters of the training algorithms,\emph{
we emphasize that we adopt exactly the same configuration for both
our model and the baseline. }Specifically, we perform training for
100 epochs, as also suggested in \cite{n2n-mem}. Adagrad is carried
out by splitting our training data into 32 minibatches. The learning
rate is initialized at $\eta=0.01$, and is annealed every 25 epochs
by $\eta/2$, until the maximum number of epochs is reached (similar
to \cite{n2n-mem}). Glorot initialization for all trained models
is performed via a Gaussian distribution with zero mean and $\sigma=0.1$.
The postulated models employ an external memory size of 50 sentences.
Nil words are padded with zero embedding (zero one-hot encodings).
The embedding space size, $\delta$, is set to 20; this is shown in
\cite{n2n-mem} to work best for the MemN2N model. In order to calculate
the output (predicted answer) for each problem, 3 computational steps
(\emph{hops}) are performed, similar to the suggestions of \cite{n2n-mem}. 

\subsection{Quantitative Assessment}

In this Section, the accuracy of the model-generated predictions is
measured and reported. In order for a trained model to be considered
successful in some type of task, we stipulate that a 95\% accuracy
must be reached, similar to \cite{n2n-mem}. The test-set accuracy
results obtained under our prescribed experimental setup are provided
in Table II. Expectably, increasing the number of MC samples drawn
to perform inference improves performance in the most challenging
of the task types, while retaining performance in the rest. As we
observe, our approach manages to pass our set success threshold of
95\% accuracy in 9 task types (\# 1, 4, 6, 11, 12, 13, 14, 15, 20),
thus outperforming the MemN2N baseline which passes the threshold
in only 5 cases (\# 1, 4, 12, 15, 20). Another characteristic finding
is that our approach outperforms the baseline in most tasks, and achieves
the same performance in the few rest. In our perception, this finding
vouches for the better capacity of our approach to learn the underlying
distributions in the modeled dataset. Of course, one also observes
that our method does not offer significant improvements on the types
of tasks for which the baseline has low accuracy. However, this apparently
constitutes an inherent weakness of the whole learning paradigm adopted
by MEM-NN networks; this cannot be rectified by introducing better
inference mechanisms, as we do in this work.

Further, to show how the test error of the considered approaches converges
over the training algorithm epochs, in Fig. 3 we depict the evolution
of the test error for an indicative task type, in one execution of
our experiments. We observe that both variants of our approach (i.e.,
using one or ten MC samples for performing inference) converge gradually
and consistently over the training algorithm epochs. In contrast,
the baseline MemN2N approach appears to reach its best performance
early-on during training, and subsequently remains almost stable. 

\subsection{Qualitative Assessment }

To provide some qualitative insights into the inferred question-answering
rationale of our approach, and how this compares to the original MemN2N,
in Tables III - V we illustrate what the inferred attention vectors
look like in three indicative test cases. More specifically, we record
which fact each model mostly focuses its attention on, on every hop;
further, we compare this result to the supporting facts included in
the dataset. Our so-obtained results indicate that the proposed approach
manages to better focus on the most salient information (sentences),
as indicated by the provided supporting facts. This outcome offers
a strong intuitive explanation of the reasons why $t$-MEM-NN appears
to outperform the baseline, in most of the considered task types.

\subsection{Computational Times}

Apart from inferential accuracy, the computational costs of a devised
method constitute another aspect which affects its efficacy. To allow
for objectively examining this aspect, we have developed all the evaluated
algorithms using the same software platform, and executed them on
the same machine (each time without concurrently running any other
user application).\footnote{Our source codes have been developed in Python, using the TensorFlow
library \cite{tensorflow2015-whitepaper}. We run our experiments
on an Intel Xeon server with 64GB RAM and an NVIDIA Tesla K40 GPU.} Then, we recorded the resulting wall-clock times, for both model
training and inference. 

As we have observed, baseline MemN2N training requires an average
of 146.5 msec per minibatch, while our approach imposes a negligible
increase, requiring an average of 149.8 msec. This is reasonable,
since training of our model entails the same set of computations as
baseline MemN2N, with the only exception being the computation of
the $t$-divergence terms pertaining to the degrees of freedom parameters,
which are of linear complexity. Thus the observed slight increase
in computational times, which is clearly worth it for the improved
model performance.

Turning to the computational costs of the inference algorithm, we
observe that our approach requires computational times comparable
to MemN2N in order to generate one answer. This is clearly reasonable,
since both models entail the same set of feedforward computations.
On the other hand, it is significant to underline that the extra computational
costs of $t$-MEM-NN that arise from an increase in the number of
MC samples drawn to perform inference (from just one to ten) are completely
negligible. Indeed, an average increase of 0.1 msec was observed.
This was well-expected, since the extra matrix multiplications that
arise from the use of multiple drawn MC samples are completely parallelizable
over commercially available, modern GPU hardware.

\subsection{Computational Complexity}

Let us denote as $n$ the dimensionality of the model input. In essence,
this corresponds to the size of the used dictionary, $V$, as discussed
in Section II.A. In both the cases of the MemN2N and $t$-MEM-NN models,
forward propagation is dominated by the same type of matrix multiplications.
In the case of MemN2N, these are effected by making use of the model
parameter estimators, $\boldsymbol{\Theta}$; in the case of the $t$-MEM-NN
counterpart, we use samples of these parameters drawn by making use
of their corresponding means, $\boldsymbol{\mu}_{\boldsymbol{\Theta}}$,
diagonal covariances, $\boldsymbol{\sigma}_{\boldsymbol{\Theta}}^{2}$
, and degrees of freedom, $\nu_{\boldsymbol{\Theta}}$. On the other
hand, based on our model description provided in Section II.A, a MemN2N
or $t$-MEM-NN model comprising $k$ hops in memory comprises $2k+1$
layers. Then, following \cite{complexity}, and considering that each
model layer may comprise $n$ output units at most (in which case
it extracts overcomplete representations), the run-time complexity
of the model inference algorithm (forward propagation) becomes $\mathcal{O}(n^{3}(2k+1))=\mathcal{O}(n^{3})$.
This is common for the MemN2N and $t$-MEM-NN models. 

Studying the case of backprop training of the MemN2N and $t$-MEM-NN
models is more involved. Specifically, by examining Eq. (24) we observe
that the objective function involved in $t$-MEM-NN model training
entails one extra operation compared to the simple categorical cross-entropy
of the MemN2N model; this is the set of $\Gamma(\cdot)$ functions,
computed over all the degrees of freedom parameters, $\nu_{\boldsymbol{\Theta}}$.
Our computation of the $\Gamma(\cdot)$ function is based on the Lanczos
approximation \cite{lanczos}; this essentially reduces to a simple
vector inner product and some additional elementary computations.
Similarly we approximate its derivative, widely known as the Digamma
function. Thus, the dominant source of computationl costs of model
training, for both models, are identical. Considering for simplicity
that the total number of training algorithm iterations is $\mathcal{O}(n)$,
this yields a computational complexity of $\mathcal{O}(n^{5}(2k+1))=\mathcal{O}(n^{5})$
for both MemN2N and $t$-MEM-NN \cite{complexity}.

\subsection{Further Insights}

\subsubsection{Effect of the number of hops}

In the previous experimental evaluations, we performed three memory
hops, following the suggestions of \cite{n2n-mem}. Yet, it is desirable
to know how $t$-MEM-NN performance may be affected if we change this
number. To get a proper answer to this question, we repeat our experiments
considering only one memory hop, as well as an increased number of
five hops. We provide the so-obtained results in Tables VI and VII.
As we observe, conducting only one hop in memory results in a significant
performance impairment in the vast majority of the considered task
types. This way, our model manages to pass the success threshold in
only two of the considered task types (\#1 and 12), as opposed to
the eight task types attained when performing three memory hops. On
the other hand, a further increase of the number of hops from three
to five seems to undermine the obtained performance. Indeed, $t$-MEM-NN
passes the success threshold in only six task types, namely task types
\#1, 4, 11, 12, 15, 20. We posit that these outcomes are due to the
structure of the used dataset, as it also becomes obvious from the
number of available supporting facts, which is more than one in most
cases, but it seldom exceeds three. 

\begin{table}
\caption{$t$-MEM-NN accuracy in the test set, performing just one memory hop.}

\centering{}%
\begin{tabular}{|l||c||c|}
\hline 
\multicolumn{3}{|c|}{Test Accuracy (\%)}\tabularnewline
\hline 
\hline 
Task type & 1 sample & 10 samples\tabularnewline
\hline 
1: 1 supporting fact  & 100 & 100\tabularnewline
2: 2 supporting facts  & 34 & \textbf{35}\tabularnewline
3: 3 supporting facts & 23 & 23\tabularnewline
4: 2 argument relations & 79 & \textbf{80}\tabularnewline
5: 3 argument relations & 87 & 87\tabularnewline
6: yes/no questions & 69 & 69\tabularnewline
7: counting & 52 & \textbf{56}\tabularnewline
8: lists/sets & 33 & \textbf{35}\tabularnewline
9: simple negation & 77 & \textbf{80}\tabularnewline
10: indefinite knowledge & 44 & \textbf{47}\tabularnewline
11: basic coreference & 25 & \textbf{28}\tabularnewline
12: conjunction & 96 & \textbf{99}\tabularnewline
13: compound coreference & 92 & \textbf{93}\tabularnewline
14: time reasoning & 21 & \textbf{23}\tabularnewline
15: basic deduction & 57 & \textbf{75}\tabularnewline
16: basic induction & 44 & \textbf{46}\tabularnewline
17: positional reasoning & 48 & 48\tabularnewline
18: size reasoning & 85 & \textbf{86}\tabularnewline
19: path finding & 9 & 9\tabularnewline
20: agent’s motivation & 83 & \textbf{84}\tabularnewline
\hline 
\end{tabular}
\end{table}

\begin{table}
\caption{$t$-MEM-NN accuracy in the test set, increasing the number of memory
hops to five.}

\centering{}%
\begin{tabular}{|l||c||c|}
\hline 
\multicolumn{3}{|c|}{Test Accuracy (\%)}\tabularnewline
\hline 
\hline 
Task type & 1 sample & 10 samples\tabularnewline
\hline 
1: 1 supporting fact  & 100 & 100\tabularnewline
2: 2 supporting facts  & 78 & \textbf{82}\tabularnewline
3: 3 supporting facts & 56 & \textbf{57}\tabularnewline
4: 2 argument relations & 93 & \textbf{96}\tabularnewline
5: 3 argument relations & 87 & 87\tabularnewline
6: yes/no questions & 73 & \textbf{78}\tabularnewline
7: counting & 78 & \textbf{81}\tabularnewline
8: lists/sets & 87 & \textbf{89}\tabularnewline
9: simple negation & 83 & \textbf{86}\tabularnewline
10: indefinite knowledge & 82 & \textbf{84}\tabularnewline
11: basic coreference & 96 & \textbf{97}\tabularnewline
12: conjunction & 98 & \textbf{99}\tabularnewline
13: compound coreference & 90 & 90\tabularnewline
14: time reasoning & 87 & \textbf{89}\tabularnewline
15: basic deduction & 94 & \textbf{96}\tabularnewline
16: basic induction & 44 & \textbf{46}\tabularnewline
17: positional reasoning & 50 & \textbf{51}\tabularnewline
18: size reasoning & 89 & \textbf{90}\tabularnewline
19: path finding & 11 & \textbf{14}\tabularnewline
20: agent’s motivation & 100 & 100\tabularnewline
\hline 
\end{tabular}
\end{table}

\begin{table}
\caption{$t$-MEM-NN accuracy in the test set, reducing the embedding size
to $\delta=10$.}

\centering{}%
\begin{tabular}{|l||c||c|}
\hline 
\multicolumn{3}{|c|}{Test Accuracy (\%)}\tabularnewline
\hline 
\hline 
Task type & 1 sample & 10 samples\tabularnewline
\hline 
1: 1 supporting fact  & 100 & 100\tabularnewline
2: 2 supporting facts  & 55 & \textbf{56}\tabularnewline
3: 3 supporting facts & 31 & \textbf{34}\tabularnewline
4: 2 argument relations & 79 & \textbf{81}\tabularnewline
5: 3 argument relations & 83 & 83\tabularnewline
6: yes/no questions & 60 & \textbf{61}\tabularnewline
7: counting & 77 & 77\tabularnewline
8: lists/sets & 85 & 85\tabularnewline
9: simple negation & 70 & 70\tabularnewline
10: indefinite knowledge & 71 & \textbf{75}\tabularnewline
11: basic coreference & 96 & \textbf{97}\tabularnewline
12: conjunction & 98 & 98\tabularnewline
13: compound coreference & 92 & 92\tabularnewline
14: time reasoning & 83 & 83\tabularnewline
15: basic deduction & 70 & \textbf{74}\tabularnewline
16: basic induction & 45 & 45\tabularnewline
17: positional reasoning & 50 & 50\tabularnewline
18: size reasoning & 87 & 87\tabularnewline
19: path finding & 10 & 10\tabularnewline
20: agent’s motivation & 100 & 100\tabularnewline
\hline 
\end{tabular}
\end{table}

\begin{table}
\caption{$t$-MEM-NN accuracy in the test set, increasing the embedding size
to $\delta=50$.}

\centering{}%
\begin{tabular}{|l||c||c|}
\hline 
\multicolumn{3}{|c|}{Test Accuracy (\%)}\tabularnewline
\hline 
\hline 
Task type & 1 sample & 10 samples\tabularnewline
\hline 
1: 1 supporting fact  & 100 & 100\tabularnewline
2: 2 supporting facts  & 73 & \textbf{78}\tabularnewline
3: 3 supporting facts & 52 & \textbf{55}\tabularnewline
4: 2 argument relations & 90 & \textbf{91}\tabularnewline
5: 3 argument relations & 85 & \textbf{86}\tabularnewline
6: yes/no questions & 72 & \textbf{78}\tabularnewline
7: counting & 79 & \textbf{81}\tabularnewline
8: lists/sets & 86 & \textbf{89}\tabularnewline
9: simple negation & 86 & \textbf{87}\tabularnewline
10: indefinite knowledge & 80 & \textbf{82}\tabularnewline
11: basic coreference & 95 & \textbf{96}\tabularnewline
12: conjunction & 99 & 99\tabularnewline
13: compound coreference & 93 & \textbf{95}\tabularnewline
14: time reasoning & 81 & \textbf{84}\tabularnewline
15: basic deduction & 98 & \textbf{100}\tabularnewline
16: basic induction & 44 & \textbf{45}\tabularnewline
17: positional reasoning & 50 & \textbf{52}\tabularnewline
18: size reasoning & 89 & \textbf{91}\tabularnewline
19: path finding & 11 & \textbf{13}\tabularnewline
20: agent’s motivation & 100 & 100\tabularnewline
\hline 
\end{tabular}
\end{table}

\subsubsection{Altering the embedding space dimensionality}

In addition, it is interesting to examine what the effect of the embedding
space dimensionality is on the performance of our approach. Indeed,
it is reasonable to expect that the larger the embedding space the
more potent a postulated model is. However, the entailed increase
in the trainable model parameters does also come at the cost of considerably
higher overfitting tendencies. These might eventually undermine the
obtained accuracy profile of $t$-MEM-NN.

To examine these aspects, we repeat our experiments considering a
smaller embeddings size than the one suggested in \cite{n2n-mem},
specifically $\delta=10$, as well as a much larger one, specifically
$\delta=50$. The outcomes of this investigation are depicted in Tables
VIII and IX, respectively. As we observe, decreasing the postulated
embedding space dimensionality to $\delta=10$ results in worse model
performance, since the model passes the success threshold in only
four task types (\#1, 11, 12, 20). Similarly interesting are the findings
pertaining to an increase of the embedding space size to $\delta=50$.
In this case, average model performance over all the considered task
types remains essentially stable. Thus, it seems that model performance
reaches a plateau as we continue to increase the size of the embeddings.
Note also that postulating either $\delta=10$ or $\delta=50$ results
in $t$-MEM-NN passing the success threshold in exactly the same set
of task types as when we postulate $\delta=20$.

To summarize, increasing the embedding space size does not appear
to be worth the extra computational costs. Indeed, $t$-MEM-NN requires
an extra 2.2 msec to generate one answer when $\delta$ increases
from 20 to 50, which represents an average increase by 62\%. On the
other hand, predictive accuracy does not yield any considerable increase.

\subsubsection{Joint task modeling}

In all the previous experiments, we have trained a distinct model
on each one of the 20 types of QA tasks included in the considered
\emph{en-1K} bAbI benchmark. However, one could also consider jointly
training one single model on data from all the included task types.\emph{
}Clearly, one may argue that this alternative approach might make
it more difficult for the trained model to distinguish between fine
patterns. However, it is also the case that, by training a joint model
on all task types, we also allow for a significantly reduced overfitting
tendency (by increasing the effective number of training data). Hence,\emph{
}we consider training a single model on all the task types; we train
for 60 epochs, and anneal the learning rate every 15 epochs. 

Our results, obtained by setting the latent space dimensionality equal
to $\delta=20$, and by employing 3 memory hops, are depicted in Table
X. It is evident that, similar to the single-task setup, inference
using 10 MC samples yields better average performance than using just
one. Considering the set success threshold of 95\% accuracy, we obtain
that our method succeeds in 9 task types (\# 1, 6, 9, 10, 12, 13,
14, 15, 20); this way, it outperforms baseline MemN2N by one task.
Note also that $t$-MEM-NN performance is greater than MemN2N in all
these tasks.

\begin{table}
\caption{Joint-modeling setup: Accuracy results in the test set.}

\centering{}%
\begin{tabular}{|l||c||c||c|}
\hline 
\multicolumn{4}{|c|}{Test Accuracy (\%)}\tabularnewline
\hline 
\hline 
 & Baseline & \multicolumn{2}{c|}{$t$-MEM-NN}\tabularnewline
\hline 
Task type & MemN2N & 1 sample & 10 samples\tabularnewline
\hline 
1: 1 supporting fact  & 99 & 100  & \textbf{100}\tabularnewline
2: 1 supporting facts  & \textbf{86} & 61 & 79\tabularnewline
3: 3 supporting facts & \textbf{71} & 45 & 60\tabularnewline
4: 2 argument relations & 85 & 77 & 85\tabularnewline
5: 3 argument relations & 86 & 82 & 86\tabularnewline
6: yes/no questions & 96 & 99 & \textbf{100}\tabularnewline
7: counting & 84 & 84 & \textbf{85}\tabularnewline
8: lists/sets & 89 & 88 & 89\tabularnewline
9: simple negation & 96 & 99 & \textbf{99}\tabularnewline
10: indefinite knowledge & 92 & 91 & \textbf{95}\tabularnewline
11: basic coreference & \textbf{94} & 90 & 91\tabularnewline
12: conjunction & 98 & 99 & \textbf{100}\tabularnewline
13: compound coreference & 97 & 93 & \textbf{98}\tabularnewline
14: time reasoning & 88 & 91 & \textbf{96}\tabularnewline
15: basic deduction & 98 & 97 & \textbf{100}\tabularnewline
16: basic induction & 46 & 46 & \textbf{48}\tabularnewline
17: positional reasoning & 55 & 55 & \textbf{58}\tabularnewline
18: size reasoning & 59 & 67 & \textbf{71}\tabularnewline
19: path finding & 10 & 14 & \textbf{17}\tabularnewline
20: agent’s motivation & 100 & 100 & 100\tabularnewline
\hline 
\end{tabular}
\end{table}

\subsubsection{Experimental evaluation with the rest of the available tasks}

Further, for completeness sake, we examine how the performance of
our model compares to the competition when it comes to considering
the rest of the tasks available in the bAbI dataset. That is, we report
results on the 20 QA tasks that are developed in the Hindi language,
that comprise both 1K as well as 10K training examples, or employ
random shuffling. These are denoted as \emph{en-10K, hn-1K, hn-10K,
shuffle-1K, }and \emph{shuffle-10K, }respectively\emph{. }Our findings,
obtained by setting the latent space dimensionality equal to $\delta=20$,
and by employing 3 memory hops, are depicted in Tables XI-XV. As we
observe, in all cases our approach exceeds the 95\% success threshold
in more tasks than the baseline. This is yet another result that vouches
for the validity of our theoretical claims, and the efficacy of our
algorithmic construction and derivations. 

\begin{table}
\caption{Accuracy results in the \textbf{Hindi/1k} test set.}

\centering{}%
\begin{tabular}{|l||c||c||c|}
\hline 
\multicolumn{4}{|c|}{Test Accuracy (\%)}\tabularnewline
\hline 
\hline 
 & Baseline & \multicolumn{2}{c|}{$t$-MEM-NN}\tabularnewline
\hline 
Task type & MemN2N & 1 sample & 10 samples\tabularnewline
\hline 
1: 1 supporting fact  & 99 & 100  & \textbf{100}\tabularnewline
2: 1 supporting facts  & 85 & 85 & \textbf{88}\tabularnewline
3: 3 supporting facts & 53 & 50 & \textbf{55}\tabularnewline
4: 2 argument relations & 97 & 96 & \textbf{98}\tabularnewline
5: 3 argument relations & 88 & 85 & \textbf{89}\tabularnewline
6: yes/no questions & 89 & 89 & 89\tabularnewline
7: counting & 83 & 83 & \textbf{84}\tabularnewline
8: lists/sets & 88 & 87 & \textbf{90}\tabularnewline
9: simple negation & 89 & 89 & \textbf{92}\tabularnewline
10: indefinite knowledge & 78 & 75 & \textbf{82}\tabularnewline
11: basic coreference & 84 & 90 & \textbf{98}\tabularnewline
12: conjunction & 99 & 99 & 99\tabularnewline
13: compound coreference & 89 & 93 & \textbf{95}\tabularnewline
14: time reasoning & 93 & 92 & \textbf{96}\tabularnewline
15: basic deduction & 100 & 98 & 100\tabularnewline
16: basic induction & 45 & 45 & \textbf{47}\tabularnewline
17: positional reasoning & 49 & 45 & \textbf{51}\tabularnewline
18: size reasoning & 86 & 86 & \textbf{93}\tabularnewline
19: path finding & 13 & 12 & 13\tabularnewline
20: agent’s motivation & 100 & 100 & 100\tabularnewline
\hline 
\end{tabular}
\end{table}

\begin{table}
\caption{Accuracy results in the \textbf{Shuffled/1k} test set.}

\centering{}%
\begin{tabular}{|l||c||c||c|}
\hline 
\multicolumn{4}{|c|}{Test Accuracy (\%)}\tabularnewline
\hline 
\hline 
 & Baseline & \multicolumn{2}{c|}{$t$-MEM-NN}\tabularnewline
\hline 
Task type & MemN2N & 1 sample & 10 samples\tabularnewline
\hline 
1: 1 supporting fact  & 100 & 100  & 100\tabularnewline
2: 1 supporting facts  & 79 & 75 & 79\tabularnewline
3: 3 supporting facts & 49 & 50 & \textbf{54}\tabularnewline
4: 2 argument relations & 95 & 93 & 95\tabularnewline
5: 3 argument relations & 86 & 85 & 86\tabularnewline
6: yes/no questions & 91 & 90 & \textbf{94}\tabularnewline
7: counting & 80 & 80 & \textbf{84}\tabularnewline
8: lists/sets & 87 & 87 & \textbf{89}\tabularnewline
9: simple negation & 89 & 89 & \textbf{94}\tabularnewline
10: indefinite knowledge & 79 & 80 & \textbf{83}\tabularnewline
11: basic coreference & 83 & 92 & \textbf{100}\tabularnewline
12: conjunction & 99 & 99 & \textbf{100}\tabularnewline
13: compound coreference & 89 & 90 & \textbf{95}\tabularnewline
14: time reasoning & 91 & 91 & \textbf{96}\tabularnewline
15: basic deduction & 100 & 99 & 100\tabularnewline
16: basic induction & 45 & 45 & \textbf{46}\tabularnewline
17: positional reasoning & 50 & 49 & \textbf{52}\tabularnewline
18: size reasoning & 86 & 88 & \textbf{91}\tabularnewline
19: path finding & 12 & 12 & \textbf{13}\tabularnewline
20: agent’s motivation & 100 & 100 & 100\tabularnewline
\hline 
\end{tabular}
\end{table}

\begin{table}
\caption{Accuracy results in the \textbf{en/10k} test set.}

\centering{}%
\begin{tabular}{|l||c||c||c|}
\hline 
\multicolumn{4}{|c|}{Test Accuracy (\%)}\tabularnewline
\hline 
\hline 
 & Baseline & \multicolumn{2}{c|}{$t$-MEM-NN}\tabularnewline
\hline 
Task type & MemN2N & 1 sample & 10 samples\tabularnewline
\hline 
1: 1 supporting fact  & 100 & 100  & 100\tabularnewline
2: 1 supporting facts  & 98 & 97 & \textbf{100}\tabularnewline
3: 3 supporting facts & 83 & 85 & \textbf{89}\tabularnewline
4: 2 argument relations & 100 & 98 & 100\tabularnewline
5: 3 argument relations & 99 & 99 & 99\tabularnewline
6: yes/no questions & 100 & 99 & 100\tabularnewline
7: counting & 95 & 95 & \textbf{97}\tabularnewline
8: lists/sets & 97 & 98 & \textbf{100}\tabularnewline
9: simple negation & 99 & 98 & 99\tabularnewline
10: indefinite knowledge & 96 & 96 & \textbf{99}\tabularnewline
11: basic coreference & 91 & 94 & \textbf{100}\tabularnewline
12: conjunction & 100 & 100 & 100\tabularnewline
13: compound coreference & 94 & 94 & \textbf{98}\tabularnewline
14: time reasoning & 97 & 95 & \textbf{100}\tabularnewline
15: basic deduction & 100 & 100 & 100\tabularnewline
16: basic induction & 47 & 47 & 47\tabularnewline
17: positional reasoning & 57 & 57 & 57\tabularnewline
18: size reasoning & 89 & 88 & \textbf{92}\tabularnewline
19: path finding & 33 & 33 & \textbf{36}\tabularnewline
20: agent’s motivation & 100 & 100 & 100\tabularnewline
\hline 
\end{tabular}
\end{table}

\begin{table}
\caption{Accuracy results in the \textbf{Hindi/10k} test set.}

\centering{}%
\begin{tabular}{|l||c||c||c|}
\hline 
\multicolumn{4}{|c|}{Test Accuracy (\%)}\tabularnewline
\hline 
\hline 
 & Baseline & \multicolumn{2}{c|}{$t$-MEM-NN}\tabularnewline
\hline 
Task type & MemN2N & 1 sample & 10 samples\tabularnewline
\hline 
1: 1 supporting fact  & 100 & 100  & 100\tabularnewline
2: 1 supporting facts  & 97 & 97 & \textbf{100}\tabularnewline
3: 3 supporting facts & 83 & 81 & \textbf{91}\tabularnewline
4: 2 argument relations & 99 & 99 & 99\tabularnewline
5: 3 argument relations & 98 & 98 & \textbf{99}\tabularnewline
6: yes/no questions & 100 & 100 & 100\tabularnewline
7: counting & 94 & 93 & \textbf{97}\tabularnewline
8: lists/sets & 97 & 97 & \textbf{99}\tabularnewline
9: simple negation & 98 & 98 & \textbf{99}\tabularnewline
10: indefinite knowledge & 93 & 93 & \textbf{97}\tabularnewline
11: basic coreference & 94 & 96 & \textbf{100}\tabularnewline
12: conjunction & 100 & 100 & 100\tabularnewline
13: compound coreference & 96 & 97 & \textbf{100}\tabularnewline
14: time reasoning & 98 & 98 & \textbf{100}\tabularnewline
15: basic deduction & 100 & 100 & 100\tabularnewline
16: basic induction & 46 & 43 & \textbf{47}\tabularnewline
17: positional reasoning & 56 & 56 & \textbf{57}\tabularnewline
18: size reasoning & 91 & 91 & \textbf{94}\tabularnewline
19: path finding & 31 & 32 & \textbf{35}\tabularnewline
20: agent’s motivation & 100 & 100 & 100\tabularnewline
\hline 
\end{tabular}
\end{table}

\begin{table}
\caption{Accuracy results in the \textbf{Shuffled/10k} test set.}

\centering{}%
\begin{tabular}{|l||c||c||c|}
\hline 
\multicolumn{4}{|c|}{Test Accuracy (\%)}\tabularnewline
\hline 
\hline 
 & Baseline & \multicolumn{2}{c|}{$t$-MEM-NN}\tabularnewline
\hline 
Task type & MemN2N & 1 sample & 10 samples\tabularnewline
\hline 
1: 1 supporting fact  & 100 & 100  & 100\tabularnewline
2: 1 supporting facts  & 97 & 98 & \textbf{100}\tabularnewline
3: 3 supporting facts & 83 & 83 & \textbf{88}\tabularnewline
4: 2 argument relations & 100 & 100 & 100\tabularnewline
5: 3 argument relations & 99 & 98 & 99\tabularnewline
6: yes/no questions & 100 & 100 & 100\tabularnewline
7: counting & 95 & 95 & \textbf{97}\tabularnewline
8: lists/sets & 96 & 96 & \textbf{99}\tabularnewline
9: simple negation & 99 & 99 & \textbf{100}\tabularnewline
10: indefinite knowledge & 97 & 97 & \textbf{99}\tabularnewline
11: basic coreference & 92 & 94 & \textbf{100}\tabularnewline
12: conjunction & 100 & 100 & 100\tabularnewline
13: compound coreference & 96 & 96 & \textbf{100}\tabularnewline
14: time reasoning & 98 & 95 & \textbf{100}\tabularnewline
15: basic deduction & 100 & 100 & 100\tabularnewline
16: basic induction & 47 & 47 & \textbf{48}\tabularnewline
17: positional reasoning & 56 & 56 & \textbf{57}\tabularnewline
18: size reasoning & 90 & 90 & \textbf{92}\tabularnewline
19: path finding & 35 & 35 & \textbf{38}\tabularnewline
20: agent’s motivation & 100 & 100 & 100\tabularnewline
\hline 
\end{tabular}
\end{table}

\subsubsection{Do we actually need to infer heavy-tailed posteriors?}

As we have already discussed, the central assumption in the formulation
of our model that the imposed posteriors are of multivariate Student's-$t$
form allows to account for heavy-tailed underlying densities, with
power-law nature. However, a question that naturally arises is whether
this assumption actually addresses an existing problem. That is, whether
the underlying densities are actually heavy-tailed. To address this
question, we can leverage some attractive properties of the Student's-$t$
distribution. Specifically, as we have explained in Section II.C,
the degrees of freedom parameter of a Student's-$t$ density controls
how heavy its tails are. This way, a model employing Student's-$t$
densities effectively modifies, through model fitting, how heavy its
tails are, to account for the actual needs of the application at hand. 

Therefore, examining the degrees of freedom values of the fitted model
posteriors is a natural means of deducing whether the imposition of
Student's-$t$ densities is actually worthwhile, or a simpler Gaussian
assumption would suffice. Our findings can be summarized as follows:
In all the experimental cases reported above, the posteriors over
the input embedding matrices $\boldsymbol{A}$ and $\boldsymbol{B}$,
given in (19) and (20), yield values $\nu_{\boldsymbol{A}},\nu_{\boldsymbol{B}}\leq5$,
while for the output embeddings $\boldsymbol{C}$, given by (21),
we have $\nu_{\boldsymbol{C}}\leq20$. These findings imply that all
our fitted models end up requiring degrees of freedom parameter values
low enough to account for quite heavy tails. Thus, our empirical experimental
findings vouch for the efficacy of our assumptions.

\section{Guess the Number}

We conclude our experimental investigations by considering a setup
that allows for us to evaluate whether our model is capable of learning
latent abstract concepts by engaging in conversation with a teacher.
Specifically, our devised experimental setup emulates a kid's game
named ``Guess the number.'' This well-known game is played by two
entities, the teacher and the student; on each round, the teacher
picks an integer number between given boundaries, and the student
tries to guess which number the teacher has originally selected. When
the student guesses a number different than the target, the teacher
provides them a hint whether the guessed number is greater or less
than it. On the sequel, the student has to make another guess, following
the limits dictated in the preceding conversation (i.e., all previous
guesses and provided hints). The game continues until either the student
guesses the target number or we reach the maximum allowed numbers
of tries.

Under this experimental rationale, we have constructed datasets that
correspond to two different scenarios; in these, the chosen numbers
lie between: (a) 0 and 10; and (b) 0 and 100. The maximum number of
tries is set to 100 for both scenarios. In addition, we perform evaluation
with a diverse number of training examples including 100, 1K, and
10K, in order to assess the effect of the training dataset size. The
latent space dimensionality of all the evaluated models is set equal
to $\delta=20$, while we employ 3 memory hops, similar to Section
IV.B. Model evaluation is performed on 100 distinct test games, in
all cases. Inference for $t$-MEM-NN is run with the number of drawn
MC samples set to 1 or 10; training is performed by drawing just one
MC sample.

For the purpose of quantitative performance evaluation of the trained
models, we have defined and use three metrics: (i) \emph{accuracy},
which describes the average percentage of correct decisions; a guess
is considered correct when the guessed number is within the limits
defined by the conversation's history; (ii) \emph{success}, which
describes the average percentage of games where the target number
was correctly guessed within the preset limit of 100 tries; and (iii)
\emph{rounds}, the average number of guesses made before the target
was found. The last metric obviously concerns only successful games. 

Our so-obtained results are depicted in Tables XVI and XVII. To allow
for the reader to get an insight into the construction of the considered
game, as well as the generated outputs of MemN2N and our proposed
approach, we provide two characteristic output samples of the evaluated
models in Table XVIII. According to the outcome of this assessment,
it is obvious that our proposed approach outperforms the baseline
model in all metrics for all scenarios. 

\begin{table}
\begin{centering}
\caption{Guess the number: selected numbers take values between 0 and 10.}
\begin{tabular}{|l||c||c||c|}
\hline 
\multicolumn{4}{|c|}{100 Training examples}\tabularnewline
\hline 
\hline 
 & Baseline & $t$-MEM-NN & \multicolumn{1}{c|}{$t$-MEM-NN}\tabularnewline
\hline 
Metric & MemN2N & 1 sample & 10 samples\tabularnewline
\hline 
Accuracy (\%) & 77 & 87 & \textbf{91}\tabularnewline
Success (\%) & 86 & 95 & \textbf{98}\tabularnewline
Rounds & 4.3 & 3.9 & \textbf{3.6}\tabularnewline
\hline 
\end{tabular}\\
\par\end{centering}
\begin{centering}
\begin{tabular}{|l||c||c||c|}
\hline 
\multicolumn{4}{|c|}{1k Training examples}\tabularnewline
\hline 
\hline 
 & Baseline & $t$-MEM-NN & \multicolumn{1}{c|}{$t$-MEM-NN}\tabularnewline
\hline 
Metric & MemN2N & 1 sample & 10 samples\tabularnewline
\hline 
Accuracy (\%) & 99 & 100 & \textbf{100}\tabularnewline
Success (\%) & 100 & 100 & \textbf{100}\tabularnewline
Rounds & 4.2 & 3.8 & \textbf{3.6}\tabularnewline
\hline 
\end{tabular}\\
\par\end{centering}
\begin{centering}
\begin{tabular}{|l||c||c||c|}
\hline 
\multicolumn{4}{|c|}{10k Training examples}\tabularnewline
\hline 
\hline 
 & Baseline & $t$-MEM-NN & \multicolumn{1}{c|}{$t$-MEM-NN}\tabularnewline
\hline 
Metric & MemN2N & 1 sample & 10 samples\tabularnewline
\hline 
Accuracy (\%) & 97 & 99 & \textbf{100}\tabularnewline
Success (\%) & 98 & 100 & \textbf{100}\tabularnewline
Rounds & 4.0 & 3.8 & \textbf{3.7}\tabularnewline
\hline 
\end{tabular}\\
\par\end{centering}
\centering{}\label{guess10}
\end{table}

\begin{table}
\begin{centering}
\caption{Guess the number: selected numbers take values between 0 and 100.}
\begin{tabular}{|l||c|c||c|}
\hline 
\multicolumn{4}{|c|}{100 Training examples}\tabularnewline
\hline 
\hline 
 & Baseline & $t$-MEM-NN & \multicolumn{1}{c|}{$t$-MEM-NN}\tabularnewline
\hline 
Metric & MemN2N & 1 sample & 10 samples\tabularnewline
\hline 
Accuracy (\%) & 17 & 19 & \textbf{21}\tabularnewline
Success (\%) & 20 & 28 & \textbf{31}\tabularnewline
Rounds & 7.7 & 5.1 & \textbf{3.2}\tabularnewline
\hline 
\end{tabular}\\
\par\end{centering}
\begin{centering}
\begin{tabular}{|l||c||c||c|}
\hline 
\multicolumn{4}{|c|}{1k Training examples}\tabularnewline
\hline 
\hline 
 & Baseline & $t$-MEM-NN & \multicolumn{1}{c|}{$t$-MEM-NN}\tabularnewline
\hline 
Metric & MemN2N & 1 sample & 10 samples\tabularnewline
\hline 
Accuracy (\%) & 30 & 49 & \textbf{55}\tabularnewline
Success (\%) & 45 & 61 & \textbf{72}\tabularnewline
Rounds & 8.7 & 8.2 & \textbf{7.7}\tabularnewline
\hline 
\end{tabular}\\
\par\end{centering}
\begin{centering}
\begin{tabular}{|l||c|c||c|}
\hline 
\multicolumn{4}{|c|}{10k Training examples}\tabularnewline
\hline 
\hline 
 & Baseline & $t$-MEM-NN & \multicolumn{1}{c|}{$t$-MEM-NN}\tabularnewline
\hline 
Metric & MemN2N & 1 sample & 10 samples\tabularnewline
\hline 
Accuracy (\%) & 63 & 78 & \textbf{85}\tabularnewline
Success (\%) & 68 & 89 & \textbf{97}\tabularnewline
Rounds & 12.6 & 9.1 & \textbf{8.5}\tabularnewline
\hline 
\end{tabular}\\
\par\end{centering}
\centering{}\label{guess100}
\end{table}

\begin{table}
\begin{centering}
\caption{$t$-MEM-NN: Sample output of ``Guess the number'' game.}
\par\end{centering}
\begin{centering}
-{}-{}-{}-{}-{}-{}-{}-{}-{}-{}-{}-{}-{}-{}-{}-{}-{}-{}-{}-{}-{}-{}-{}-{}-{}-{}-{}-{}-{}-{}- 
\par\end{centering}
\begin{centering}
Model: \textbf{t-MEM-NN }
\par\end{centering}
\begin{centering}
min: \textbf{0} 
\par\end{centering}
\begin{centering}
max: \textbf{100} 
\par\end{centering}
\begin{centering}
Train examples: \textbf{1000} 
\par\end{centering}
\begin{centering}
-{}-{}-{}-{}-{}-{}-{}-{}-{}-{}-{}-{}-{}-{}-{}-{}-{}-{}-{}-{}-{}-{}-{}-{}-{}-{}-{}-{}-{}-{}-
\par\end{centering}
\begin{centering}
-{}-{}-{}-{}-{}-{}-{}-{}-{}-{}-{}-{}-{}-{}-{}-{}-{}-{}-{}-{}-{}-{}-{}-{}-{}-{}-{}-{}-{}-{}- 
\par\end{centering}
\begin{centering}
{*}{*}{*}{*} TESTING MODEL STARTS {*}{*}{*}{*} 
\par\end{centering}
\begin{centering}
-{}-{}-{}-{}-{}-{}-{}-{}-{}-{}-{}-{}-{}-{}-{}-{}-{}-{}-{}-{}-{}-{}-{}-{}-{}-{}-{}-{}-{}-{}- 
\par\end{centering}
\begin{centering}
Select a number between 0 and 100 
\par\end{centering}
\begin{centering}
Round: 1 
\par\end{centering}
\begin{centering}
-{}-{}-{}-{}-{}-{}-{}-{}-{}-{}-{}-{}-
\par\end{centering}
\begin{centering}
\textbf{Selection: 76}
\par\end{centering}
\begin{centering}
\textbf{\textcolor{green}{Correct: Selection within Bounds!}}\textcolor{green}{{} }
\par\end{centering}
\begin{centering}
Accuracy: = 1.0 
\par\end{centering}
\begin{centering}
\textbf{min\_num: -1 }
\par\end{centering}
\begin{centering}
\textbf{max\_num: 101 }
\par\end{centering}
\begin{centering}
\textbf{Hint: Target is a smaller number }
\par\end{centering}
\begin{centering}
Round: 2 
\par\end{centering}
\begin{centering}
-{}-{}-{}-{}-{}-{}-{}-{}-{}-{}-{}-{}-
\par\end{centering}
\begin{centering}
\textbf{Selection: 100}
\par\end{centering}
\begin{centering}
\textbf{\textcolor{red}{Wrong: Selection Out of Bounds! }}
\par\end{centering}
\begin{centering}
Accuracy: = 0.5
\par\end{centering}
\begin{centering}
\textbf{min\_num: -1 }
\par\end{centering}
\begin{centering}
\textbf{max\_num: 76 }
\par\end{centering}
\begin{centering}
\textbf{Hint: Target is a smaller number }
\par\end{centering}
\begin{centering}
Round: 3 
\par\end{centering}
\begin{centering}
-{}-{}-{}-{}-{}-{}-{}-{}-{}-{}-{}-{}-
\par\end{centering}
\begin{centering}
\textbf{Selection: 1}
\par\end{centering}
\begin{centering}
\textbf{\textcolor{green}{Correct: Selection within Bounds!}}\textcolor{green}{{} }
\par\end{centering}
\begin{centering}
Accuracy: = 0. 66
\par\end{centering}
\begin{centering}
\textbf{min\_num: -1 }
\par\end{centering}
\begin{centering}
\textbf{max\_num: 76 }
\par\end{centering}
\begin{centering}
\textbf{Hint: Target is a larger number }
\par\end{centering}
\begin{centering}
Round: 4 
\par\end{centering}
\begin{centering}
-{}-{}-{}-{}-{}-{}-{}-{}-{}-{}-{}-{}-
\par\end{centering}
\begin{centering}
\textbf{Selection: 59}
\par\end{centering}
\begin{centering}
\textbf{\textcolor{green}{Correct: Selection within Bounds!}}\textcolor{green}{{} }
\par\end{centering}
\begin{centering}
Accuracy: = 0.75 
\par\end{centering}
\begin{centering}
\textbf{min\_num: 1 }
\par\end{centering}
\begin{centering}
\textbf{max\_num: 76} 
\par\end{centering}
\begin{centering}
\textbf{Hint: Target is a smaller number }
\par\end{centering}
\begin{centering}
Round: 5 
\par\end{centering}
\begin{centering}
-{}-{}-{}-{}-{}-{}-{}-{}-{}-{}-{}-{}-
\par\end{centering}
\begin{centering}
\textbf{Selection: 11}
\par\end{centering}
\begin{centering}
\textbf{\textcolor{green}{Correct: Selection within Bounds!}}\textcolor{green}{{} }
\par\end{centering}
\begin{centering}
Accuracy: = 0.8 
\par\end{centering}
\begin{centering}
\textbf{min\_num: 1 }
\par\end{centering}
\begin{centering}
\textbf{max\_num: 59 }
\par\end{centering}
\begin{centering}
\textbf{Hint: Target is a larger number }
\par\end{centering}
\begin{centering}
Round: 6 
\par\end{centering}
\begin{centering}
-{}-{}-{}-{}-{}-{}-{}-{}-{}-{}-{}-{}-
\par\end{centering}
\begin{centering}
\textbf{Selection: 56}
\par\end{centering}
\begin{centering}
\textbf{\textcolor{green}{Correct: Selection within Bounds!}}\textcolor{green}{{} }
\par\end{centering}
\begin{centering}
Accuracy: = 0.83 
\par\end{centering}
\begin{centering}
\textbf{min\_num: 11 }
\par\end{centering}
\begin{centering}
\textbf{max\_num: 59} 
\par\end{centering}
\begin{centering}
\textbf{Hint: Target is a larger number }
\par\end{centering}
\begin{centering}
Round: 7 
\par\end{centering}
\begin{centering}
-{}-{}-{}-{}-{}-{}-{}-{}-{}-{}-{}-{}-
\par\end{centering}
\begin{centering}
\textbf{Selection: 54}
\par\end{centering}
\begin{centering}
\textbf{\textcolor{red}{Wrong: Selection Out of Bounds! }}
\par\end{centering}
\begin{centering}
Accuracy: = 0.71
\par\end{centering}
\begin{centering}
\textbf{min\_num: 56 }
\par\end{centering}
\begin{centering}
\textbf{max\_num: 59 }
\par\end{centering}
\begin{centering}
\textbf{Hint: Target is a larger number }
\par\end{centering}
\begin{centering}
Round: 8 
\par\end{centering}
\begin{centering}
-{}-{}-{}-{}-{}-{}-{}-{}-{}-{}-{}-{}-
\par\end{centering}
\begin{centering}
\textbf{Selection: 57}
\par\end{centering}
\begin{centering}
\textbf{\textcolor{green}{Correct: Selection within Bounds!}}\textcolor{green}{{} }
\par\end{centering}
\begin{centering}
Accuracy: = 0.75 
\par\end{centering}
\begin{centering}
\textbf{min\_num: 56 }
\par\end{centering}
\begin{centering}
\textbf{max\_num: 59 }
\par\end{centering}
\begin{centering}
{*}{*}{*}{*}{*}{*}{*}{*}{*}{*}{*}{*}{*}{*}{*}{*}{*}{*}{*}{*}{*}{*}{*}{*}{*}{*}{*}{*}{*}{*}{*}{*}{*}{*}{*}{*}{*}{*}{*}{*}{*}{*}{*}{*}{*}{*}{*}{*}{*}
\par\end{centering}
\begin{centering}
\textbf{Congratulations, the target is 57 }
\par\end{centering}
\begin{centering}
\textbf{You found the correct answer after 8 rounds }
\par\end{centering}
\centering{}\textbf{Accuracy: 0.75}
\end{table}

\section{Conclusions }

In this paper, we attacked the problem of modeling long-term dependencies
in sequential data. Specifically, we focused on question-answering
bots; these inherently require the ability to perform multiple computational
steps of analyzing observed patterns over long temporal horizons,
and on multiple time-scales. To achieve this goal, one may resort
to the paradigm of neural attention models that operate over large
external memory modules. This is a recent development in the field
of machine learning, yielding state-of-the-art performance in challenging
benchmark tasks. 

In this context, the core contribution of our work was the provision
of a novel inferential framework for this class of models, which allows
to account for the uncertainty in the modeled data. This is a significant
issue when dealing with sparse datasets, which are prevalent in real-world
the considered tasks. In addition, our method was carefully crafted
so as to best accommodate data with heavy-tailed distributions, which
are typical in multivariate sequences. 

To achieve these goals, we devised a novel Bayesian inference-driven
algorithmic formulation of end-to-end-trainable MEM-NNs. Specifically,
we considered a stochastic model formulation, where the trainable
parameters (embedding matrices) of the network are imposed appropriate
prior distributions, and corresponding posteriors are inferred by
means of variational Bayes. To allow for accommodating heavy-tailed
data, we postulated latent variables belonging to the $t$-exponential
family; specifically, we considered multivariate Student's-$t$ densities.
In the same vein, and in order to allow for reaping the most out of
the data modeling power of Student's-$t$ densities, we performed
variational inference for our model under a novel objective function
construction. This was based on a $t$-divergence criterion, which
offers an attractive alternative to the KL divergence (that is minimized
in conventional variational Bayes), tailored to heavy-tailed data. 

We performed an extensive experimental evaluation of our approach
using challenging question-answering benchmarks. We provided thorough
insights into the inferential outcomes of our approach, and how these
compare to the competition. We also illustrated that our proposed
approach achieves the reported accuracy improvement without undermining
computational efficiency, both in training time and in prediction
generation time.

One research direction that we have not considered in this work concerns
the possibility of imposing nonelliptical or skewed distributions
on the postulated latent variables. Indeed, many researchers in the
past have shown that conventional generative models for sequential
data, e.g. hidden Markov models, can yield significant benefits by
considering nonelliptically contoured latent state densities, such
as the multivariate normal inverse Gaussian (MNIG) distribution \cite{pami10}.
On the other hand, the efficacy and the potential advantages of introducing
skewed latent variable assumptions in the context of DL models was
empirically demonstrated in \cite{asydgm}. Nevertheless, such assumptions
certainly come at the cost of increased computational complexity.
Hence, we reckon that progressing beyond the elliptical class of distributions
for formulating the assumptions of our model is a worthwhile future
research direction. It might allow for even higher modeling performance,
but requires novel theoretical developments to ensure retainment of
the method's computational efficiency. Thus, these opportunities remain
to be explored in our future research.

\section*{Acknowledgment}

We gratefully acknowledge the support of NVIDIA Corporation with the
donation of one Tesla K40 GPU used for this research.

\bibliographystyle{IEEEtran}
\bibliography{t-mem}

\begin{thebibliography}{10}
\providecommand{\url}[1]{#1}
\csname url@samestyle\endcsname
\providecommand{\newblock}{\relax}
\providecommand{\bibinfo}[2]{#2}
\providecommand{\BIBentrySTDinterwordspacing}{\spaceskip=0pt\relax}
\providecommand{\BIBentryALTinterwordstretchfactor}{4}
\providecommand{\BIBentryALTinterwordspacing}{\spaceskip=\fontdimen2\font plus
\BIBentryALTinterwordstretchfactor\fontdimen3\font minus
  \fontdimen4\font\relax}
\providecommand{\BIBforeignlanguage}[2]{{%
\expandafter\ifx\csname l@#1\endcsname\relax
\typeout{** WARNING: IEEEtran.bst: No hyphenation pattern has been}%
\typeout{** loaded for the language `#1'. Using the pattern for}%
\typeout{** the default language instead.}%
\else
\language=\csname l@#1\endcsname
\fi
#2}}
\providecommand{\BIBdecl}{\relax}
\BIBdecl

\bibitem{turing}
A.~Graves, G.~Wayne, and I.~Danihelka, ``{Neural Turing machines},'' in
  \emph{Proc. NIPS}, 2014.

\bibitem{mem}
J.~Weston, S.~Chopra, and A.~Bordes, ``Memory networks,'' in \emph{Proc. ICLR},
  2015.

\bibitem{n2n-mem}
S.~Sukhbaatar, A.~Szlam, J.~Weston, and R.~Fergus, ``End-to-end memory
  networks,'' in \emph{Proc. NIPS}, 2015.

\bibitem{sum}
A.~Rush, S.~Chopra, and J.~Weston, ``A neural attention model for abstractive
  sentence summarization,'' in \emph{Proc. ACL}, 2015.

\bibitem{zipf}
G.~K. Zipf, \emph{The psychology of language}.\hskip 1em plus 0.5em minus
  0.4em\relax Houghton-Mifflin, 1935.

\bibitem{aevb}
D.~Kingma and M.~Welling, ``Auto-encoding variational {Bayes},'' in \emph{Proc.
  ICLR'14}, 2014.

\bibitem{aevb2}
D.~P. Kingma, D.~J. Rezende, S.~Mohamed, and M.~Welling, ``Semi-supervised
  learning with deep generative models,'' in \emph{Proc. NIPS'14}, 2014.

\bibitem{aevb3}
D.~J. Rezende, S.~Mohamed, and D.~Wierstra, ``Stochastic backpropagation and
  approximate inference in deep generative models,'' in \emph{Proc. ICML},
  2014.

\bibitem{aevb4}
D.~J. Rezende and S.~Mohamed, ``Variational inference with normalizing flows,''
  in \emph{Proc. ICML}, 2015.

\bibitem{vb-weights}
C.~Blundell, J.~Cornebise, K.~Kavukcuoglu, and D.~Wierstra, ``Weight
  uncertainty in neural networks,'' in \emph{Proc. ICML}, 2015.

\bibitem{vbg}
M.~Jordan, Z.~Ghahramani, T.~Jaakkola, and L.~Saul, ``An introduction to
  variational methods for graphical models,'' in \emph{Learning in Graphical
  Models}, M.~Jordan, Ed.\hskip 1em plus 0.5em minus 0.4em\relax Dordrecht:
  Kluwer, 1998, pp. 105--162.

\bibitem{entropy}
M.~J. Wainwright and M.~I. Jordan, ``Graphical models, exponential families,
  and variational inference,'' \emph{Foundations and Trends in Machine
  Learning}, vol.~1, no. 1-2, pp. 1--305, 2008.

\bibitem{key-30}
A.~Kosinski, ``A procedure for the detection of multivariate outliers,''
  \emph{Computational Statistics and Data Analysis}, vol.~29, pp. 145--161,
  1999.

\bibitem{asydgm}
H.~Partaourides and S.~P. Chatzis, ``Asymmetric deep generative models,''
  \emph{Neurocomputing}, vol. 241, pp. 90--96, 2017.

\bibitem{tsalis}
C.~Tsallis, ``Possible generalization of {Boltzmann-Gibbs} statistics,''
  \emph{J. Stat. Phys.}, vol.~52, pp. 479--487, 1998.

\bibitem{tsalis2}
A.~Sousa and C.~Tsallis, ``Student's $t$- and $r$-distributions: Unified
  derivation from an entropic variational principle,'' \emph{Physica A}, vol.
  236, pp. 52--57, 1994.

\bibitem{tsalis3}
C.~Tsallis, R.~S. Mendes, and A.~R. Plastino, ``The role of constraints within
  generalized nonextensive statistics.'' \emph{Physica A}, vol. 261, pp.
  534--554, 1998.

\bibitem{naudts}
J.~Naudts, ``Deformed exponentials and logarithms in generalized
  thermostatistics,'' \emph{Physica A}, vol. 316, pp. 323-- 334, 2002.

\bibitem{naudts2}
------, ``Generalized thermostatistics and mean-field theory,'' \emph{Physica
  A}, vol. 332, pp. 279--300, 2004.

\bibitem{naudts3}
------, ``Estimators, escort proabilities, and $\phi$-exponential families in
  statistical physics,'' \emph{Journal of Inequalities in Pure and Applied
  Mathematics}, vol.~5, no.~4, 2004.

\bibitem{vbshmm}
S.~P. Chatzis and D.~Kosmopoulos, ``{A Variational Bayesian Methodology for
  Hidden Markov Models utilizing Student's-t Mixtures},'' \emph{Pattern
  Recognition}, vol.~44, no.~2, pp. 295--306, Feb. 2011.

\bibitem{vbtmfa}
S.~Chatzis, D.~Kosmopoulos, and T.~Varvarigou, ``Signal modeling and
  classification using a robust latent space model based on $t$
  distributions,'' \emph{IEEE Trans. Signal Processing}, vol.~56, no.~3, March
  2008.

\bibitem{shmm}
------, ``Robust sequential data modeling using an outlier tolerant hidden
  {Markov} model,'' \emph{IEEE Trans. Pattern Analysis and Machine
  Intelligence}, vol.~31, no.~9, pp. 1657--1669, 2009.

\bibitem{key-19}
G.~McLachlan and D.~Peel, \emph{Finite Mixture Models}.\hskip 1em plus 0.5em
  minus 0.4em\relax New York: Wiley Series in Probability and Statistics, 2000.

\bibitem{tVB}
N.~Ding, S.~N. Vishwanathan, and Y.~Qi, ``$t$-divergence based approximate
  inference,'' in \emph{Proc. NIPS}, 2011.

\bibitem{attias}
H.~Attias, ``A variational {Bayesian} framework for graphical models,'' in
  \emph{Proc. NIPS'00}, 2000.

\bibitem{liu}
C.~Liu and D.~Rubin, ``{ML} estimation of the $t$ distribution using {EM} and
  its extensions, {ECM} and {ECME},'' \emph{Statistica Sinica}, vol.~5, no.~1,
  pp. 19--39, 1995.

\bibitem{gsm}
D.~Andrews and C.~Mallows, ``Scale mixtures of normal distributions,'' \emph{J.
  Royal Stat. Soc. B}, vol.~36, pp. 99--102, 1974.

\bibitem{vbsmm2}
M.~Svens\'en and C.~M. Bishop, ``Robust {Bayesian} mixture modelling,''
  \emph{Neurocomputing}, vol.~64, pp. 235--252, 2005.

\bibitem{adagrad}
J.~Duchi, E.~Hazan, and Y.~Singer, ``Adaptive subgradient methods for online
  learning and stochastic optimization,'' \emph{JMLR}, vol.~12, pp. 2121--
  2159, 2010.

\bibitem{glorot}
X.~Glorot and Y.~Bengio, ``Understanding the difficulty of training deep
  feedforward neural networks,'' in \emph{Proc. AISTATS}, 2010.

\bibitem{babi}
J.~Weston, A.~Bordes, S.~Chopra, and T.~Mikolov, ``{Towards AI-complete
  question answering: A set of prerequisite toy tasks},'' in \emph{Proc. ICLR},
  2016.

\bibitem{tensorflow2015-whitepaper}
\BIBentryALTinterwordspacing
M.~Abadi, A.~Agarwal, P.~Barham, E.~Brevdo, Z.~Chen, C.~Citro, G.~S. Corrado,
  A.~Davis, J.~Dean, M.~Devin, S.~Ghemawat, I.~Goodfellow, A.~Harp, G.~Irving,
  M.~Isard, Y.~Jia, R.~Jozefowicz, L.~Kaiser, M.~Kudlur, J.~Levenberg,
  D.~Man\'{e}, R.~Monga, S.~Moore, D.~Murray, C.~Olah, M.~Schuster, J.~Shlens,
  B.~Steiner, I.~Sutskever, K.~Talwar, P.~Tucker, V.~Vanhoucke, V.~Vasudevan,
  F.~Vi\'{e}gas, O.~Vinyals, P.~Warden, M.~Wattenberg, M.~Wicke, Y.~Yu, and
  X.~Zheng, ``{TensorFlow}: Large-scale machine learning on heterogeneous
  systems,'' 2015, software available from tensorflow.org. [Online]. Available:
  \url{http://tensorflow.org/}
\BIBentrySTDinterwordspacing

\bibitem{complexity}
\BIBentryALTinterwordspacing
 [Online]. Available:
  \url{https://kasperfred.com/posts/computational-complexity-of-neural-networks}
\BIBentrySTDinterwordspacing

\bibitem{lanczos}
G.~R. Pugh, ``An analysis of the {Lanczos} gamma approximation,'' Ph.D.
  dissertation, Faculty of Mathematics, Department of Science, University of
  British Columbia, 2004.

\bibitem{pami10}
S.~Chatzis, ``{Hidden Markov Models with Nonelliptically Contoured State
  Densities},'' \emph{IEEE Trans. Pattern Analysis and Machine Intelligence},
  vol.~32, no.~12, pp. 2297--2304, Dec. 2010.

\end{thebibliography}

\end{document}